\documentclass[10pt,twocolumn,letterpaper]{article}

\usepackage[]{cvpr} %

\usepackage{color}

\newcommand{\bb}{\boldsymbol}
\newcommand{\ul}{\underline}

\usepackage{amsmath,amsfonts,bm}

\def\eqref#1{equation~\ref{#1}}

\def\1{\bm{1}}

\def\vw{{\bm{w}}}
\def\vx{{\bm{x}}}

\def\vz{{\bm{z}}}

\DeclareMathAlphabet{\mathsfit}{\encodingdefault}{\sfdefault}{m}{sl}
\SetMathAlphabet{\mathsfit}{bold}{\encodingdefault}{\sfdefault}{bx}{n}

\def\sX{{\mathbb{X}}}
\def\sY{{\mathbb{Y}}}

\def\uu{\underline}

\newcommand{\E}{\mathbb{E}}

\newcommand{\R}{\mathbb{R}}

\usepackage{times}
\usepackage{epsfig}
\usepackage{graphicx}
\usepackage{amsmath}
\usepackage{amssymb}
\usepackage{multirow}
\usepackage{enumitem}
\usepackage[symbol]{footmisc}

\definecolor{cvprblue}{rgb}{0.21,0.49,0.74}
\usepackage[pagebackref,breaklinks,colorlinks,citecolor=cvprblue]{hyperref}

\def\paperID{6099} %
\def\confName{CVPR}
\def\confYear{2024}

\title{Taming the Tail in Class-Conditional GANs: Knowledge Sharing via Unconditional Training at Lower Resolutions}

\author{
Saeed Khorram$^{1}$, Mingqi Jiang$^{1}$, Mohamad Shahbazi $^{2}$, Mohamad H. Danesh$^{3}$, Li Fuxin $^{1}$
\\ 
$^{1}$%
\text{Oregon State University}
$^{2}$
\text{ETH Zürich}
$^{3}$
\text{McGill University} \\
{\tt \{khorrams, jiangmi, lif\}@oregonstate.edu} \\
{\tt mshahbazi@vision.ee.ethz.ch, mohamad.danesh@mail.mcgill.ca}
}

\begin{document}
\maketitle

\begin{abstract}

Despite extensive research on training generative adversarial networks (GANs) with limited training data, learning to generate images from long-tailed training distributions remains fairly unexplored. In the presence of imbalanced multi-class training data, GANs tend to favor classes with more samples, leading to the generation of low quality and less diverse samples in tail classes. In this study, we aim to improve the training of class-conditional GANs with long-tailed data. We propose a straightforward yet effective method for knowledge sharing, allowing tail classes to borrow from the rich information from classes with more abundant training data. 
More concretely, we propose modifications to existing class-conditional GAN architectures to ensure that the lower-resolution layers of the generator are trained entirely unconditionally while reserving class-conditional generation for the higher-resolution layers.
Experiments on several long-tail benchmarks and GAN architectures demonstrate a significant improvement over existing methods in both the diversity and fidelity of the generated images. The code is available at \url{https://github.com/khorrams/utlo}.
\end{abstract}

\vspace{-0.2in}
\section{Introduction}\label{sec:intro}

In the past few years, research on Generative Adversarial Networks (GANs) \cite{goodfellowGAN} has led to remarkable advances in generating realistic images \cite{DBLP:conf/iclr/BrockDS19,karras2020training, SauerS022StyleGAN-XL}. Conditional GANs \cite{mirza2014conditional} (cGANs) have garnered particular attention due to their ability to accept user inputs which can additionally guide the generation process. They enable a wide range of applications such as class-conditional image generation \cite{shahbazi2022collapse}, image manipulation~\cite{shen2021closedform}, image-to-image translation \cite{isola2017image}, super-resolution \cite{ledig2017photo} and text-to-image synthesis \cite{zhang2017stackgan}.

\begin{figure}[t!]
  \centering
  \includegraphics[width=0.92 \linewidth]{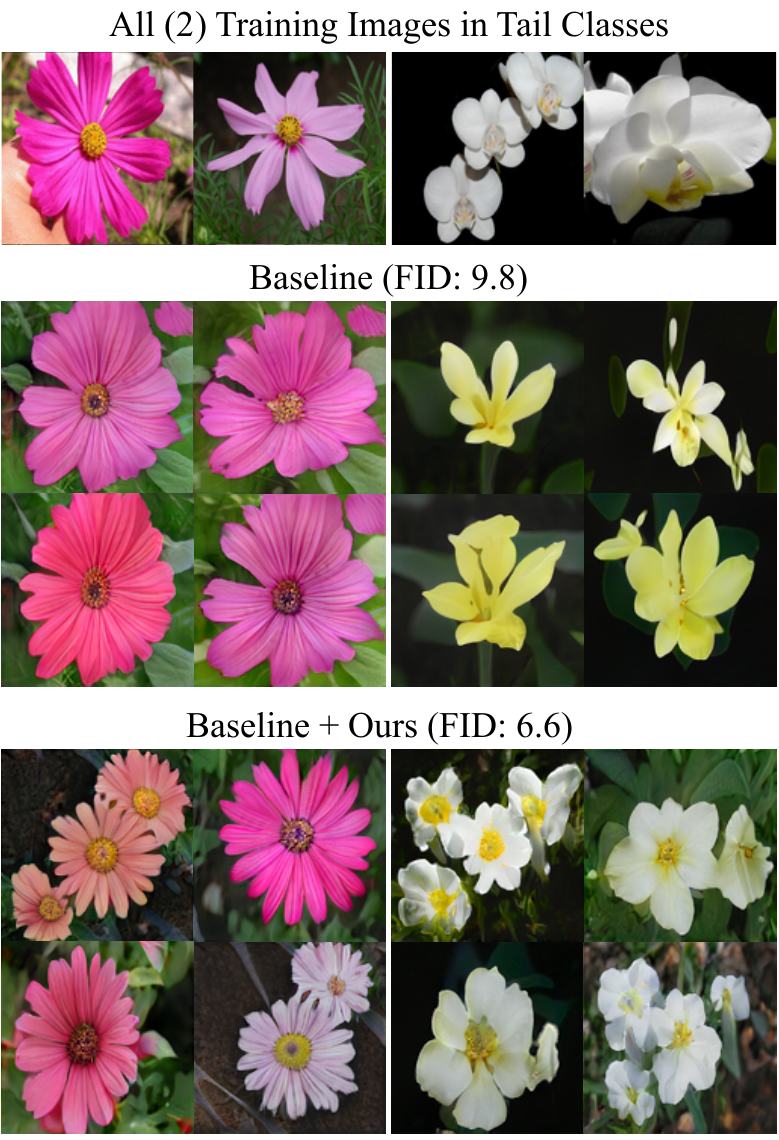}
  \vskip -0.1in
   \caption{Generating images from rare tail classes in the Flowers-LT with only \textbf{two} training images. Our proposed approach allows for a more diverse set of features such as backgrounds, colors, poses, and object layouts to be infused into the tail classes. %
   }
   \label{fig:teas}
   \vskip -0.15in
\end{figure}

Despite the advances, past research on cGANs has focused primarily on learning from balanced data. In real-world scenarios, however, data  often follows a power-law distribution, with a few classes dominating most of the training data. This is referred to as the ``\textit{long-tail (LT) problem}" \cite{zhang2021deepLT}, where most of the training data come from a few classes (referred to as the ``\textit{head}" of the distribution). In contrast, data from a large number of classes rarely occur (referred to as the ``\textit{tail}" of the distribution). This imbalance in data poses a challenge to the effective learning of tail classes, as standard machine learning algorithms tend to favor densely sampled regions of the input distribution. 

Although there has been a surge of interest in the long-tail problem in recent years, the focus has been primarily on recognition tasks \cite{yang22surveyLT}. Generative learning on the long-tail data, particularly cGANs which is the focus of this work, remains underexplored. Recently, ~\cite{rangwani2022improving, rangwani2023noisytwins} identified mode collapse on tail classes as a result of the spectral norm explosion in class-conditional BatchNorms and the collapse of the latents in the StyleGANs $\mathcal{W}$ space. While techniques such as regularization, normalization, and/or class-balancing techniques \cite{yang22surveyLT} might improve the performance of long-tail generative models to some extent, their effectiveness is limited. The primary challenge in the long-tail problem is the under-representation of tail classes in the training data, e.g., in Fig.~\ref{fig:teas} each tail class contains merely 2 training images, resulting in insufficient observation of the tail classes by the learning algorithm.

To address this challenge, we propose to leverage the abundant information in the head classes, with the aim to learn and infuse knowledge from the head classes into the tail ones, thereby enriching the training distribution for the latter. 
However, it is very difficult to explicitly disentangle the class-specific information pertaining to head classes from class-independent information shared between head and tail classes. 
We make a key observation that the similarity between head and tail instances is often \textit{higher at lower resolutions}: discriminative features such as shape or specific texture are usually unveiled at higher resolutions. Conversely, information from the lower resolutions tends to be more class-independent such as background, configuration, or direction of objects (Fig.~\ref{fig:cf10_uc}), thus can be shared between the head and tail classes.

Building upon this observation, we propose Unconditional Training at LOwer resolution (UTLO), a novel approach for training cGANs in the long-tail setup. In addition to the standard conditional GAN objective, UTLO trains the intermediate low-resolution output of the generator with an \textit{unconditional} GAN objective. The unconditional GAN objective encourages the learning of low-resolution features common to both the head and tail classes, which are infused into the subsequent layers of the generator, particularly benefiting the under-represented tail classes.

Through extensive experiments and analysis on several long-tailed datasets, we demonstrate that using UTLO to combine high-resolution conditional and low-resolution unconditional training effectively facilitates knowledge sharing between head and tail classes, thereby improving the overall generative modeling of cGANs in the long-tail setup. Due to the strong class imbalance in the long-tailed data, naive usage of existing GAN metrics can be misleading. To mitigate this issue, we propose a few practices to adapt commonly-used GAN metrics to the long-tail setup. %

Below, we highlight our main contributions:

\begin{itemize}[topsep=0pt, partopsep=0pt, itemsep=0pt, parsep=0pt]
    
    \item 
    We propose UTLO, a novel knowledge-sharing framework tailored for training cGANs in the long-tailed setup. UTLO allows infusing information from the head classes to the tail classes via an additional unconditional objective applied to the low-resolution part of the cGAN generator. 
    To the best of our knowledge, this work is the first to demonstrate that not all layers in a cGAN need to receive class-conditional information, i.e., a cGAN featuring a partially unconditional generator. %

    \item 
    We present a set of practices and metrics designed to adapt the commonly-used GAN evaluation metrics for long-tail setups, enabling a more precise evaluation of the image generation quality.
    
    \item 
    Through extensive experiments across multiple benchmarks and architectures, we validate the effectiveness of our proposed method in improving the training of cGANs in the long-tail setup, achieving state-of-the-art results across several long-tail datasets. 
\end{itemize}

\section{Related Work}

\noindent \textbf{Conditional Generative Adversarial Networks}
The initial work on conditional image generation using GANs \cite{goodfellowGAN} was presented by \cite{mirza2014conditional}, in which the class conditions were concatenated to the inputs of the generator and discriminator networks. The AC-GAN \cite{OdenaOS17ACGAN} proposed the use of auxiliary classification in the discriminator. \cite{DBLP:conf/iclr/BrockDS19} introduced BigGAN and set a milestone in large-scale and high-resolution conditional image generation. The authors of StyleGAN \cite{stylegan2Karras, karras2020training} first map the input noise and class condition to a latent style space, which is then passed to the multiple layers of the generator for image synthesis. More recently, an extension of the StyleGAN called StyleGAN-XL \cite{SauerS022StyleGAN-XL} has become the state-of-the-art in conditional image generation on several datasets and outperforms more complex and time-consuming approaches such as diffusion models \cite{DBLP:conf/nips/DhariwalN21}.

\noindent \textbf{GAN Regularization under Limited Data} Training GANs under limited data is challenging as the discriminator can memorize the training samples, resulting in the collapse of the training or quality degradation of the generated images. Recently, data augmentation and regularization techniques have been incorporated into GAN training as a means to mitigate this problem \cite{zhang2019consistency, karras2020training, zhao2020differentiable, KarmaliPARJ0B22, zhao2020image, tran2021data}. Additionally, \cite{tseng2021regularizing} introduced a regularization term to the GAN objective, which tracks the predictions of the discriminator for real and generated images using separate moving averages. \cite{kumari2022ensembling} employed off-the-shelf vision models in an ensemble of discriminators, demonstrating improved performance in both limited-data and large-scale GAN training. 

In the context of few-shot image generation, FastGAN \cite{liu2021towards} proposed a lightweight architecture with a self-supervised discriminator. \cite{Ojha_2021_CVPR} introduced cross-domain distance consistency in order to transfer diversity from a source domain to a target domain. Subsequently, \cite{KongKHK22smoothing} proposed a latent-mixup strategy to smooth the latent space through controlled latent interpolation.

Recently, \cite{zhao2020differentiable, shahbazi2022collapse} have observed that class-conditioning can cause mode collapse in the limited data regime. Their proposed work learns from limited but balanced/unlabeled data, which is different from our setup: heavily imbalanced long-tail data. \cite{shahbazi2022collapse} proposed Transitional-cGAN, a training strategy that starts with unconditional GAN objective and injects class-conditional information during a transition period. This approach was found to be effective in mitigating mode-collapse in the limited data regime. 
Our main idea of separating pathways between low-resolution and high-resolution is novel w.r.t.~\cite{shahbazi2022collapse}: to promote knowledge sharing, we modify the cGAN architecture so that the lower-resolution part of the generator is entirely unconditional while only the higher-resolution part is conditional and received class-conditional information.

\noindent \textbf{Long-tail Recognition} 
To address the long-tail recognition problem \cite{zhang2021deepLT, yang22surveyLT}, previous research has primarily focused on techniques such as class re-balancing \cite{NEURIPS2019_621461af, NEURIPS2020_2ba61cc3, menon2021longtail}, learning algorithm and model design \cite{DBLP:conf/iclr/KangXRYGFK20, Kang2021ExploringBF, DBLP:conf/cvpr/ZhongC0J21}, and information augmentation \cite{liu2020deep, NEURIPS2020_1091660f, DBLP:conf/iclr/WangLM0Y21}.
More related to our work, re-sampling \cite{zhang2021learning, guo2021long} has been widely used to handle class imbalance problems. This involves balancing the number of samples used during training through over-sampling and/or under-sampling. Oversampling increases the frequency of tail classes \cite{han2005borderline, mullick2019generative} while under-sampling reduces the imbalance by reducing the frequency of head class instances \cite{liu2020under, yao2021improved}. Note that our research diverges from prior studies on long-tail recognition by addressing the more challenging task of generative modeling of the long-tail data, as opposed to the conventional long-tail classification task.

\noindent \textbf{Training GANs on Long-tail Data} Recent years have seen a growing interest in learning from long-tail data. However, most  focus has been directed toward recognition tasks, with a limited number of studies addressing the development of generative models for long-tail data. GAMO \cite{mullick2019generative} uses adversarial training to over-sample from minority classes. \cite{pmlr-v161-rangwani21a} introduced class-balancing regularization to the GAN objective, which leverages the predictions of a pretrained classifier to improve the generation of underrepresented classes in an unconditional setting. However, this method trains an unconditional GAN and is restricted to having access to a pre-trained classifier on the long-tail data to guide the training. The most closely related work to ours is the Group Spectral Regularization (GSR) \cite{rangwani2022improving} and NoisyTwins \cite{rangwani2023noisytwins} regularizations.
The authors identified mode collapse on tail classes as a result of the spectral norm explosion in class-conditional BatchNorms and the collapse of the latents in the StyleGANs $\mathcal{W}$ space, respectively. In contrast, we present a novel approach for sharing knowledge between head and tail classes to enhance long-tail learning, which is orthogonal to regularization techniques. In addition, while \cite{rangwani2023noisytwins} is restricted to StyleGANs, our framework can be extended to different GAN architectures.

\section{Taming the Tail in cGANs} \label{sec:method}

\subsection{Background: Conditional Generative Adversarial Networks}\label{sec:bg}
Conditional Generative Adversarial Networks (cGANs) are suitable for many applications due to providing user control over the generated samples at inference time. This is particularly helpful in the long-tail setup, where it is desirable to explicitly generate samples from rare (tail) classes. In contrast, unconditional GANs are likely to generate samples that track the training distribution, resulting in most samples coming from common (head) classes. %
To train a cGAN over a dataset %
containing $n$ instances $\sX = \{\vx_1, \dots, \vx_n\}$
and their corresponding labels $\sY = \{y_1, \dots, y_n \}$ 
, we can formulate the adversarial training objective that alternatively optimizes the following loss terms,
\begin{align}
    &\mathcal{L}^D_c = \E_{\vx, y}[f_D\left(-D(\vx|y)\right)] + \E_{\vz, y}[f_D\left( D\left(G\left(\vz, y\right)\right)\right)] \notag \\
    &\mathcal{L}^G_c = \E_{\vz, y}[f_G\left(-D\left(G\left(\vz, y\right)\right)\right)]
\end{align} \label{eq:cgan}
\noindent where $G$ and $D$ are the generator and discriminator networks with their corresponding loss functions $f_G$ and $f_D$ \cite{lucic2018gans}; $\vx|y \sim p_{\text{data}}(\vx|y)$ is real data drawn from class $y$. The generator $G$ conditions on both $\vz \sim p_z(\vz)$, a random noise vector from a prior distribution over latent space, and $y$, the class conditioning vector. This adversarial training objective encourages the cGAN to reach an equilibrium between the generator and the discriminator, resulting in the generator producing realistic samples indistinguishable from real data by the discriminator.

\subsection{Caveat of Training cGANs on Long-tail Data}\label{sec:motiv}
Training GANs on long-tail data is challenging due to the inherent difficulty in modeling the rare examples that comprise the tail of the data distribution. This skewness in the data distribution hinders balanced learning across classes. In particular, the discriminator does not see enough examples from the tail of the distribution during training, which can lead to poor discriminative signals for the generator. As a result, the generator learns to generate a small subset of the possible outputs that fools the inadequately learned discriminator --  exhibiting a classic ``mode collapse" scenario. %

\begin{figure}[t!]
  \centering
  \includegraphics[width=1\linewidth]{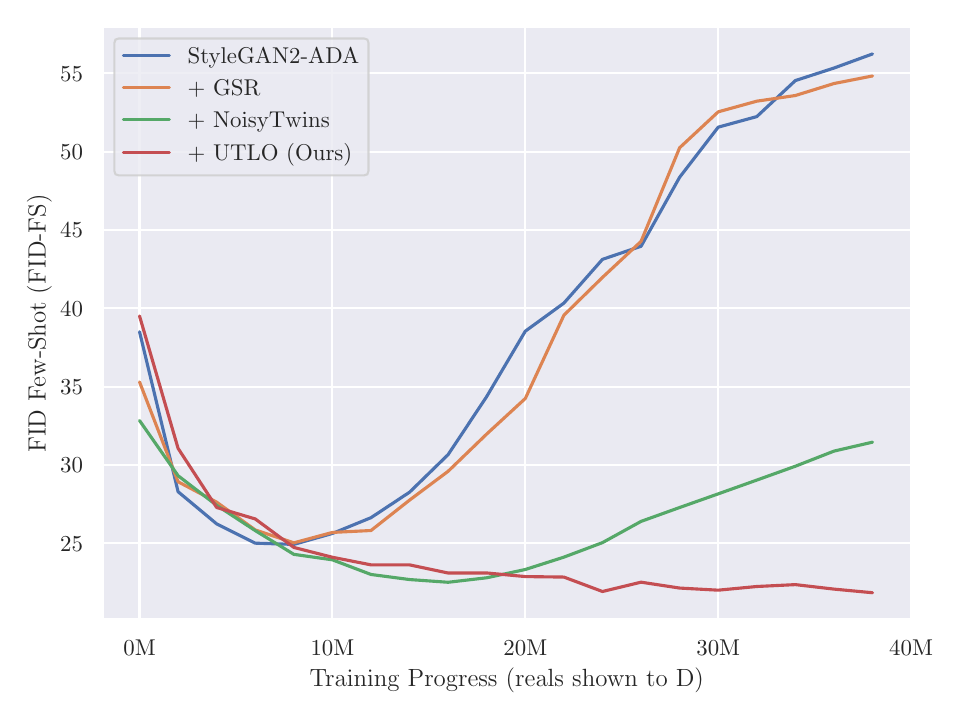}
  \includegraphics[width=1\linewidth]{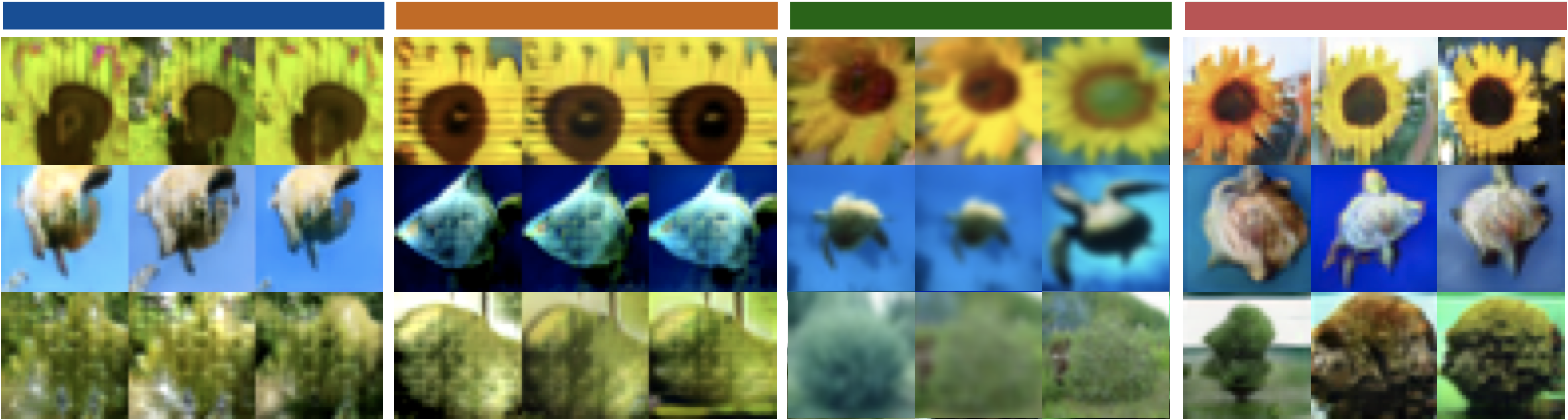}
  \vskip -0.1in
   \caption{Convergence of different methods on CIFAR100-LT ($\rho=100$), where tail classes have as few as 5 training examples. Incorporating our framework into the baseline alleviates overfitting as a result of knowledge sharing from head to rare tail classes.}
   \label{fig:overfit}
  \vskip -0.15in
\end{figure}

Although regularizations introduced in \cite{rangwani2022improving, rangwani2023noisytwins} can help mitigate the collapse of learning in tail classes that contain a sufficiently large number of samples, we still observe mode collapse occurring when a very limited number of samples are in tail classes, which is inherently challenging. 
This can be seen in Fig. \ref{fig:overfit} for tail classes of CIFAR100-LT with as few as 5 training instances. Although early-stopping can be adopted to obtain reasonable performance, the key in boosting the performance would be an approach to infuse additional diversities into the tail classes rather than only relying on techniques such as regularization. %

\begin{figure*}[t!]
  \centering
  \includegraphics[width=0.75\linewidth]{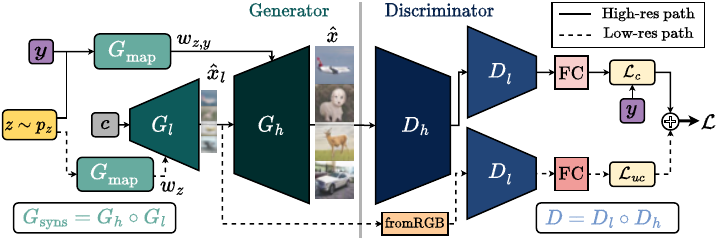}
   \caption{The proposed framework, UTLO, illustrated for the StyleGAN2-ADA architecture. Low and high resolution image pathways are used for unconditional and class-conditional objectives, respectively. $z$ is the input latent code, $y$ indicates the class embeddings, and $c$ is a constant input. Separate style vectors, $w_z$ (class-independent) and $w_{z,y}$ (class-conditional), are generated using the same z and a {shared} style-mapping network $G^{}_{\text{map}}$ which are then passed to $G_l$ and $G_h$, respectively. The high-resolution generated image $\hat{x} \in \R^{3\times H \times H}$ is passed through the discriminator $D = D_l \circ D_h$ to calculate the conditional objective $\mathcal{L}_{c}$ while the low-resolution image $\hat{x}_l \in \R^{3\times L \times L}$ is passed only through $D^{}_l$ to calculate the unconditional objective $\mathcal{L}_{uc}$. The final objective $\mathcal{L}$ is the combination of the two. While $D^{}_l$ is shared, two \textit{separate} prediction heads ($\text{FC}$ layers) are used for unconditional and conditional objectives. The $\text{fromRGB}$ is designed to increase the dimensionality of RGB channels to match the input channels of the $D^{}_{l}$.
   }
   \label{fig:uc2c}
   \vskip -0.15in
\end{figure*}

\subsection{Knowledge Sharing via Unconditional Training at Lower Resolutions (UTLO)} \label{sec:uc2c}
We devise a novel generative adversarial network (GAN) training scheme for long-tail data that aims to share knowledge from predominant \textit{head} classes to rare \textit{tail} classes, assuming some of the tail and head classes are at least \textit{coarsely} similar. %
We propose a framework that uses unconditional learning in low-resolution images / features at an intermediate layer of the generator using the \textbf{unconditional} GAN objective. This helps with learning universal features from both head and tail classes. The subsequent layers, responsible for introducing finer details at higher resolutions, are trained using a \textbf{class-conditional} GAN objective.

In our proposed framework, we perform conditional training on the high-resolution image output from the final layer of the generator network. This is built upon the features learned at lower layers through the use of an unconditional objective, inheriting features primarily from head classes. The conditioning on the class labels gives control at the inference time to explicitly generate images from tail classes, meeting our design desideratum. The combination of unconditional and conditional objectives using our proposed method enables knowledge sharing, which, in turn, improves the quality of GAN training on long-tail data.

Our framework is general and applicable to many GAN architectures. In this section, we illustrate the application of our framework on StyleGAN2 with adaptive data augmentation (ADA) \cite{karras2020training}, a state-of-the-art and solid baseline for training GANs, particularly in the limited data regime. In the following, we demonstrate the necessary modifications to the architecture of the generator and discriminator networks. Although we show the necessary modifications for StyleGAN2-ADA, our method
can be easily extended to other GAN architectures such as FastGAN \cite{liu2021towards}, %
which is also used in our experiments (see Section \ref{sec:exp}).
\vspace{-0.03in}
\subsubsection{Modifying the Generator}\label{sec:gen_mod}

In the generator design of the StyleGAN2-ADA \cite{karras2020training,stylegan2Karras}, the class-conditional information, combined with the latent vector, is first embedded in the style space $w$ using the style-mapping network $G_{\text{map}}$. The style vectors are then broadcasted across the layers of the synthesis network $G_{\text{syns}}$ in order to generate diverse images. To meet our design desideratum, we dissect the synthesis network into two sub-networks $G_{\text{syns}} = G_h \circ G_l$ where $G_l$ represents the earlier part of the network that produces intermediate features and/or images at a low resolution $L \times L$. $G_h$ on the other hand, represents the latter part of $G_{\text{syns}}$ that generates the output image at the high resolution $H \times H$ where $ H > L$. 

To block the flow of class-conditional information to the lower layers of $G_{\text{syns}}$, we generate separate $w$ vectors, one containing the class-conditional embeddings, referred to as $w_{z,y}$, while for the other one, referred to as $w_z$, class information was set to zero. Note that both vectors share the same latent variable $z$ and mapping network $G_{\text{map}}$ parameters, and the only difference is the presence of class-conditional information. The lower layers of the synthesis network $G_l$ are conditioned on $w_z$ while the subsequent layers at higher resolutions receive $w_{z,y}$ (see Fig. \ref{fig:uc2c}).

Generators typically follow a network design that gradually increases the resolution of intermediate features and/or images as the network progresses. This allows us to select a desired low-resolution, such as $8\times8$ or $16\times16$, during training. Recent generator designs often incorporate skip connections \cite{DBLP:conf/cvpr/KarnewarW20} or residual connections \cite{miyato2018spectral} to improve gradient flow. The generator of StyleGAN2-ADA uses skip connections, which explicitly generates intermediate RGB images at each layer. We take this low-resolution image as the input to the discriminator. Note that in the case of a generator design that uses residual connections, a $1 \times 1$ convolutional layer can be incorporated to convert the residual channels to RGB channels. 

In a forward pass through the generator, images at low and high resolutions can be generated simultaneously without additional overhead. By inputting a latent code $z \sim p_z(\vz)$ and a target label $y$, we can obtain an unconditional synthetic image $\hat{\vx}_l \in \R^{3\times L \times L}$ along with a conditional high-resolution synthetic image $\hat{\vx} \in \R^{3\times H \times H}$ at the output. While $\hat{\vx}$ is trained to be from class $y$, $\hat{\vx}_l$ has no class-specific constraints. At inference time, the intermediate images can be discarded.

\vspace{-0.1in}
\subsubsection{Modifying the Discriminator}\label{sec:disc_mod}
Opposite to the generator design, discriminators gradually reduce the resolution of the intermediate features as the network progresses. In order to pass images to an intermediate layer of the discriminator, we dissect it into two sub-networks. Formally, we define the discriminator as $D = D_l \circ D_h$ where $D_h$ represents the earlier part of the original network that takes the images at training resolution $\R^{3\times H \times H}$. On the other hand, $D_l$ represents the latter part of the discriminator taking a lower resolution input $\R^{C\times L \times L}$ with $C$ channels.

The default discriminator design in StyleGAN2-ADA uses residual connections. This does not allow direct passing of RGB inputs to the intermediate layers. To overcome this, we add a $1 \times 1$ convolutional layer, referred to as ``fromRGB", that increases the channels of the RGB images to match $C$, the number of input channels of $D_l$.

More concretely, $D_l$ is a shared network in both the unconditional low-resolution and class-conditional high-resolution image pathways (see Fig. \ref{fig:uc2c}). To obtain separate unconditional and conditional predictions, we use two different fully-connected (FC) layers as the final layers of $D_l$. The output of $D_h$ and the low-res image $\hat{\vx}_l$ are passed through the $D_l$ which are then used for calculating the class-conditional loss $\mathcal{L}_c$ and the unconditional loss $\mathcal{L}_{uc}$, respectively.

Note that when passing real images to $D_l$ for the unconditional loss, they are first downsized to the low-resolution $L \times L$ using bi-linear interpolation to be comparable with the ones generated from $G_l$. They are then passed through the fromRGB layer and consequently $D_l$. For the class-conditional loss, the original images are passed to $D$ without any modification.

\vspace{-0.15in}
\subsubsection{Final Training Objective}\label{sec:obj_mod}
The final objective of our framework is simply a combination of the conditional and unconditional GAN losses, using a weighting parameter $\lambda$. To put it formally we have,
\begin{align}
    \mathcal{L}^D = \mathcal{L}^D_{c} + \lambda \cdot \mathcal{L}^D_{uc}, \;
    \mathcal{L}^G = \mathcal{L}^G_{c} + \lambda \cdot \mathcal{L}^G_{uc} \label{eq:objetive}
\end{align} 
\noindent where $\mathcal{L}^D_{c}$ and $\mathcal{L}^G_{c}$ represent the class-conditional losses for the discriminator and generator, respectively (as outlined in Eq. \ref{eq:cgan}), and $\mathcal{L}^D_{uc}$ and $\mathcal{L}^G_{uc}$ represent their unconditional counterparts, where $\vx$ is sampled from $p_{\text{data}}(\vx)$ instead of $p_{\text{data}}(\vx|y)$. The overall framework of our proposed method is depicted in Fig. \ref{fig:uc2c}.

\section{Experiments} \label{sec:exp}

\subsection{Setup} \label{sub:setup}

\noindent\textbf{Datasets.}  %
We use 6 different long-tailed datasets in our experiments: CIFAR10 and CIFAR100\cite{krizhevsky2009learning}, LSUN \cite{yu15lsun}, Flowers \cite{Nilsback08}, iNaturalist2019 \cite{iNaturalist2019}, and AnimalFaces \cite{animalfaces}. This selection is intended to cover a wide spectrum of image domains, dataset sizes, resolutions, and imbalance ratios (denoted as $\rho$). $\rho$ represents the ratio of the number of training samples between the classes with the most and the least number of examples. To ensure the tail classes stay \textit{few-shot}, we keep the number of images in the smallest tail classes under 50~\cite{yang22surveyLT}. A detailed description of these datasets is provided in Supp.~\ref{supp:dataset}.

\noindent\textbf{Baselines.} %
In order to demonstrate the effectiveness of our proposed method, we use various cGAN architectures covering different generator and discriminator designs and data augmentation pipelines: 

\begin{itemize}[topsep=0pt, partopsep=0pt, itemsep=0pt, parsep=0pt]
    
    \item \textit{StyleGAN2 with Adaptive Data Augmentation (ADA)} \cite{karras2020training}. We also integrate \textit{Transitional} training \cite{shahbazi2022collapse}, \textit{Group Spectral Regularization (GSR)} \cite{rangwani2022improving}, and  \textit{NoisyTwins} \cite{rangwani2023noisytwins} to this baseline.
    
    \item \textit{Projected GAN (PGAN)} \cite{sauer2021projected}: we use projected discriminator with \textit{Differentiable Augmentation (DA)} \cite{zhao2020differentiable}. We also add GSR \cite{rangwani2022improving} to this baseline.
\end{itemize}

For comprehensive details on the baselines, hyperparameters, and implementation, please refer to Supp.~\ref{supp:impl}.
\vspace{-0.05in}
\subsection{Evaluation Metrics For Long-Tail Datasets}\label{sec:metrics}
\vspace{-0.05in}
Evaluating the images generated by GANs on a long-tailed setup poses challenges, primarily due to the imbalanced data and access to a limited number of samples for the tail classes. 
For our experiments, we employ widely-used GAN metrics: Fr\'echet Inception Distance (FID) \cite{heusel2017gans}, and Kernel Inception Distance (KID) \cite{binkowski2018demystifying}. Following ~\cite{heusel2017gans,karras2020training}, we report all metrics by computing statistics between $50k$ generated images and all available training images. In the following, we present a set of practices for adapting the aforementioned metrics to the long-tail setup.\\
\indent First, when a larger and (more) balanced dataset following the long-tail training data distribution is available, it is used for metric calculation, e.g., full CIFAR10 before artificial imbalance. For naturally imbalanced datasets such as iNaturalist2019, we use the unmodified training set to calculate metrics. Note that when generating images for evaluation, we sample from the same distribution as the available real dataset, which may be imbalanced.  While FID and KID indicate how well the real data distribution matches the generated images across all classes, we suggest additional metrics to mitigate the disproportional influence of head classes in data statistic calculations.\\
\indent In response to the varied number of training instances across classes in a long-tail setup, we propose evaluating metrics specifically on few-shot (FS) categories ~\cite{yang22surveyLT}. We calculate the FID and KID for the few-shot subset and refer to them as \textit{``FID-FS"} and \textit{``KID-FS"}, respectively. This is tailored to evaluate the quality of the generated samples in the tail classes. Unlike the standard FID/KID, we maintain an equal number of real images across all classes during our FID-FS/KID-FS calculation. This emphasizes learning quality on tail classes, irrespective of any imbalances that might be present in the few-shot subset. Reporting both standard and few-shot metrics provides a more comprehensive evaluation of long-tail learning, considering the performance of both head and tail classes.

\vspace{-0.05 in}
\subsection{Results}
In the following, we report the results obtained across different benchmarks and cGAN architectures. We pick the \textit{best} model for reporting results as the one with the lowest FID-FS from two independent runs. Note that our method exhibits significantly less reliance on early stopping compared to the baselines, as illustrated in Fig.~\ref{fig:overfit}.

\textbf{AnimalFaces-LT}  
This benchmark has a naturally occurring imbalance across its classes. We compare our method against the StyleGAN2-ADA benchmarks in terms of quantitative image quality metrics in Table.~\ref{table:af}. Our method outperforms the baselines across all metrics, as demonstrated in the results. For a visual comparison against the baselines, please refer to Supp.~\ref{supp:H}.

\begin{table}[t!]
\caption{Our proposed method, UTLO, outperforms the baselines in terms of quantitative image quality metrics on the AnimalFaces-LT dataset ($64 \times 64$ resolution).}
\vskip -0.2in

\label{table:af}
\begin{center}
\resizebox{1\linewidth}{!}{%

\begin{tabular}{l|cccc}

\multirow{2}{*}{Methods} & 
\multirow{2}{*}{FID $\downarrow$} &
\multirow{2}{*}{FID-FS $\downarrow$}  &
{KID $\downarrow$} & 
{KID-FS  $\downarrow$} \\

{} & \multicolumn{2}{c}{} & \multicolumn{2}{c}{$\times 1000$} \\

\hline \hline

{StyleGAN2-ADA UnCond. \cite{karras2020training}} 
&  {${39.4}$}&{$104.1$} & {${17.3}$}&{$27.6$} \\

\hline \hline

{StyleGAN2-ADA \cite{karras2020training}}
&  {${51.4}$}&{$87.1$} & {${24.7}$}&{$35.9$}\\

{+ Transitional~\cite{shahbazi2022collapse}}
&  {${62.1}$}&{$99.0$} & {${38.5}$}&{$48.9$} \\

{+ GSR\cite{rangwani2022improving}}              
&  {${39.2}$}&{$67.2$} & {${21.2}$}&{$32.7$}\\

{+ NoisyTwins \cite{rangwani2023noisytwins}}
&  {${29.4}$}&{$50.2$} & {${16.7}$}&{$21.2$}\\

{+ UTLO (Ours)}                    
&  {$\bb{26.2}$}&{$\bb{48.4}$} & {$\bb{12.6}$}&{$\bb{19.6}$}\\

\end{tabular}
}
\end{center}
\vskip -0.3in
\end{table}

\noindent\textbf{Ablation on the Choice of Low-resolution for Unconditional Training ($\bb{res}_{\text{uc}}$).} 
In our proposed method, one of the hyperparameter choices is to select an intermediate low-resolution $res_{\text{uc}}$ for unconditional training. All layers with equal or lower resolution than $res_{\text{uc}}$ do not receive class-conditional information. An unconditional GAN objective is added over the images and/or features at $res_{\text{uc}}$. To study the impact of $res_{\text{uc}}$, we conducted an ablation study on the AnimalFaces-LT dataset, containing images at $64 \times 64$. Table. \ref{table:af_ablation} presents the results for selecting $res_{\text{uc}}$ from resolutions lower than $64 \times 64$, namely $8 \times 8$, $16 \times 16$, and $32 \times 32$. When studying the obtained results, we observe that resolutions of $8 \times 8$ and $16 \times 16$ achieve relatively close performance. Conversely, the performance diminishes when layers up to $32 \times 32$ are trained unconditionally. This is anticipated as AnimalFaces-LT is at $64 \times 64$, leaving only one up-sampling layer to learn the class-conditional information which is shown to be insufficient. In all of our experiments, we use $res_{\text{uc}}$ of $8 \times 8$ unless otherwise stated. Additional visual analysis on the role of $res_{\text{uc}}$ as well as an ablation on the choice of unconditional training objective weight $\lambda$ (see Eq.~\ref{eq:objetive}), need for unconditional layers in the discriminator, and distinction from "coarse-to-fine" training strategies are presented in the Supp.~\ref{supp:ablation}.

\begin{table}[h!]
\vskip -0.05in
\caption{Ablation on choice of low resolution for unconditional training ($res_{\text{uc}}$).}
\vskip -0.25in

\label{table:af_ablation}
\begin{center}
\resizebox{0.7\linewidth}{!}{%

\begin{tabular}{c|cccc}

\multirow{2}{*}{$res_{\text{uc}}$} & 
\multirow{2}{*}{FID $\downarrow$} &
\multirow{2}{*}{FID-FS $\downarrow$}  &
{KID $\downarrow$} & 
{KID-FS  $\downarrow$} \\

{} & \multicolumn{2}{c}{} & \multicolumn{2}{c}{$\times 1000$} \\

\hline \hline

{$8  \times  8$}                    
&  {$\bb{26.2}$}&{$\bb{48.4}$} & {$\bb{12.6}$}&{$\bb{19.6}$}\\

{$16 \times 16$}                    
&  {${27.5}$}&{${50.3}$} & {${13.7}$}&{${20.8}$} \\

{$32 \times 32$}                    
& {${38.0}$}&{${64.9}$} & {${23.3}$}&{${34.3}$} \\

\end{tabular}
}
\end{center}
\vskip -0.25in
\end{table}

\begin{figure}[b!]
\vskip -0.15in
  \centering
  \includegraphics[width=1\linewidth]{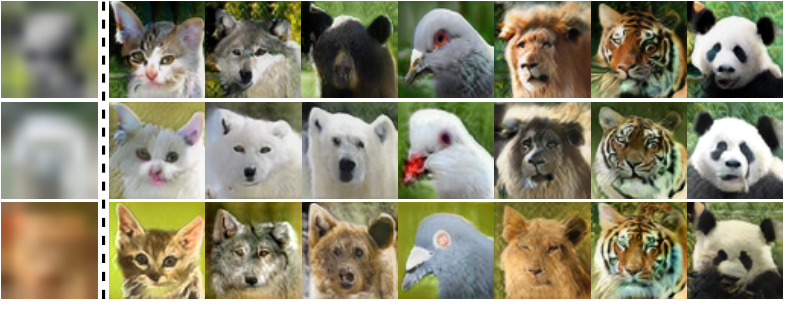}
  \vskip -0.15in
  \caption{Different class-conditional images generated given the same unconditional low-resolution images (left-most column).} %
  \label{fig:fixed_uc}
  \vskip -0.1in
\end{figure}

\begin{figure*}[t!]
  \centering
  \includegraphics[width=0.8\linewidth]{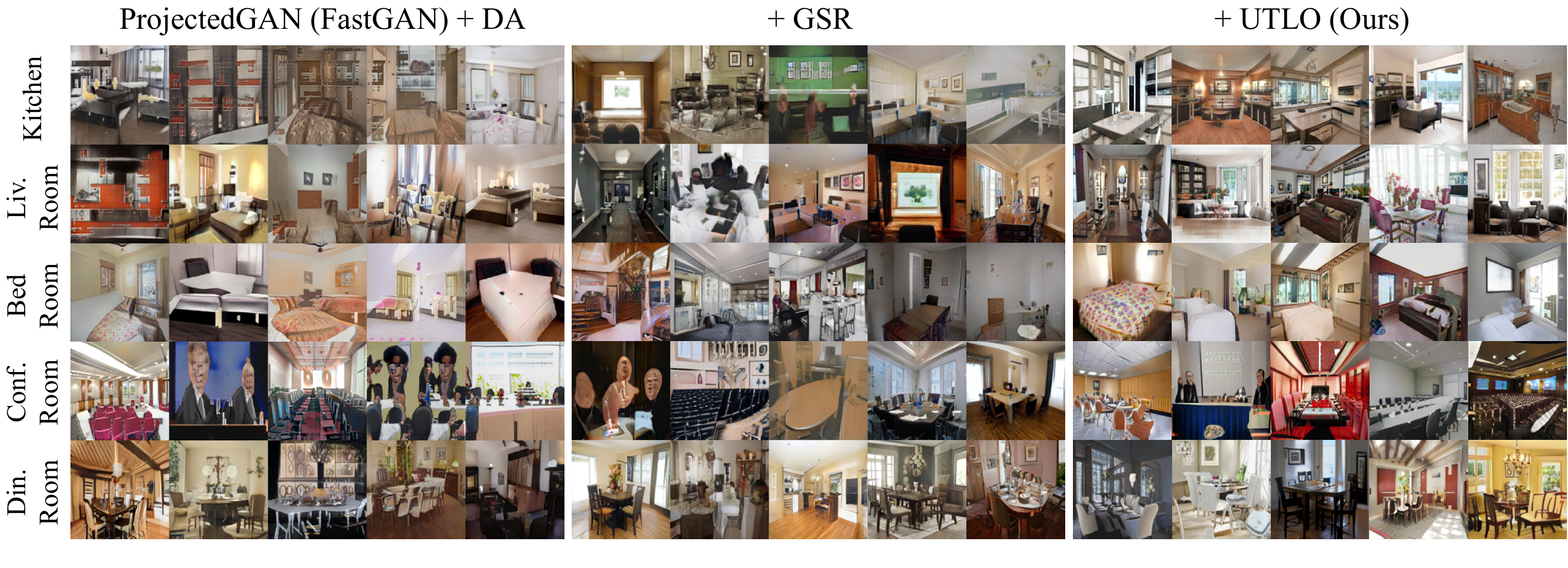}
\vskip -0.2in
   \caption{Generated images from LSUN5-LT dataset. Despite only 50 training instances for the tail class \texttt{kitchen}, the proposed UTLO framework produces diverse, high-fidelity images.}
   \label{fig:lsun}
   \vskip -0.1in
\end{figure*}

\noindent\textbf{Can any low-resolution be the starting point for high-res generations?} 
In general, our observations indicate that high-resolution images are guided, yet not entirely restricted by the low-resolution images, as shown in Fig.~\ref{fig:fixed_uc}. Each row in this figure presents a set of different high-resolution images that are generated from the same unconditional low-resolution image. This illustration reveals that high-resolution images can substantially differ from low-resolution ones, both in terms of color and texture. This suggests that subsequent conditional blocks are capable of significant modification in the background or foreground.

\noindent\textbf{Importance of FS metrics: comparison against unconditional training.}
We also include results from training an unconditional StyleGAN2-ADA model \cite{karras2020training} in Table \ref{table:af}, referred to as StyleGAN2-ADA UnCond. The unconditional model generates samples that follow the training distribution, which is mainly dominated by head classes. This bias favors unconditional learning in terms of the FID/KID metrics, which do not consider the skewness in the data distribution. Nevertheless, it can be observed that UnCond. baselines performed significantly worse in terms of few-shot metrics FID-FS/KID-FS. This suggests that relying solely on FID/KID can be misleading. We strongly recommend including few-shot metrics, i.e., FID-FS and KID-FS, when evaluating cGANs on long-tailed datasets.

\textbf{CIFAR10-LT}
In this benchmark, our method is compared with various baselines with imbalance ratios $\rho$ of $50$ and $100$. This indicates that the least frequent tail class includes only $100$ and $50$ training instances, respectively. Table. \ref{table:cifar10} quantitatively assesses the quality of the generated samples by our method and the baselines. All the baselines use the StyleGAN2-ADA \cite{karras2020training} architecture. Our method consistently outperforms the baselines across all metrics. This demonstrates that beyond improving generative learning across all classes (FID \& KID), UTLO can also notably improve the learning from tail classes (FID-FS \& KID-FS).

\noindent \textbf{Effect of Imbalance Ratio $\bb{\rho}$}. Not surprisingly, all methods demonstrate improved performance with lower values of $\rho$, as seen in Table. \ref{table:cifar10}. Further, we notice the benefits of knowledge sharing via UTLO become more pronounced as the imbalance in the dataset increases, i.e., learning tail classes becomes more challenging. It is also worth mentioning that the FID-FS and KID-FS metrics are critical to represent the quality of the few-shot classes as those are under-represented in the dataset-wide metrics that assume balanced sampling. For instance, at $\rho=50$, the FID and KID metrics obtained from UTLO show little changes, while FID-FS and KID-FS exhibit improvements close to 50\% in comparison to GSR. 

\begin{table}[t]
\caption{\small Quantitative comparison of the generated images from UTLO against baselines on CIFAR10-LT dataset with different imbalance ratios $\rho$. UTLO shows substantial improvements on few-shot (FS) metrics compared to StyleGAN2-ADA~\cite{karras2020training} while outperforming existing methods including NoisyTwins~\cite{rangwani2023noisytwins}.}
\label{table:cifar10}
\begin{center}
\vskip -0.2in
\begin{small}
\resizebox{1\linewidth}{!}{%

\begin{tabular}{c|l|cc|cc}

\multirow{2}{*}{$\rho$} & 
\multirow{2}{*}{Methods} & 
\multirow{2}{*}{FID $\downarrow$} &
\multirow{2}{*}{FID-FS $\downarrow$}  &
{KID $\downarrow$} & 
{KID-FS  $\downarrow$} \\

{} & {} & \multicolumn{2}{c|}{} & \multicolumn{2}{c}{$\times 1000$} \\
\hline \hline     

\multirow{4}{*}{50} & 
{StyleGAN2-ADA~\cite{karras2020training}}      
&  {$6.5$}&{$21.4$}&{$2.4$}&{$9.0$} \\ 
{} & {+ Transitional~\cite{shahbazi2022collapse}}
&  {${9.4}$}&{$17.7$} & {${4.7}$}&{$8.6$} \\
{} & {+ GSR~\cite{rangwani2022improving}}
&  {$6.4$}&{$21.3$}&{$\bb{2.3}$}&{$8.1$}\\
{} & {+ NoisyTwins~\cite{rangwani2023noisytwins}}
&  {${6.2}$}&{$12.2$} & {${2.4}$}&{$5.0$} \\
{} & {+ UTLO (ours)}
&  {$\bb{6.1}$}&{$\bb{11.8}$}&{$2.4$}&{$\bb{4.8}$}\\

\hline \hline

\multirow{4}{*}{100} & 
{StyleGAN2-ADA~\cite{karras2020training}  }  
&  {$9.0$}&{$24.2$} & {$4.0$}&{$9.7$} \\
{} & {+ Transitional~\cite{shahbazi2022collapse}}
&  {${11.3}$}&{$20.6$} & {${5.4}$}&{$9.2$} \\
{} & {+ GSR~\cite{rangwani2022improving}}
&  {${8.4}$}&{$24.3$} & {${3.9}$}&{$11.8$} \\
{} & {+ NoisyTwins~\cite{rangwani2023noisytwins}}
&  {${7.1}$}&{$14.1$} & {${2.9}$}&{$5.9$} \\
{} & {+ UTLO (Ours)}
&  {$\bb{6.8}$}&{$\bb{13.4}$} & {$\bb{2.8}$}&{$\bb{5.4}$} \\
\end{tabular}
}
\end{small}
\end{center}
\vskip -0.35in
\end{table}

\begin{figure}[h!]
  \centering
  \includegraphics[width=1\linewidth]{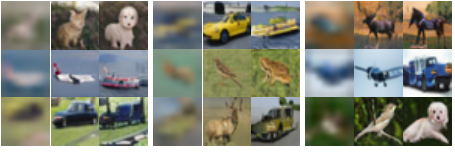}
      \vskip -0.15 in
   \caption{Knowledge sharing from head to tail classes in CIFAR10-LT dataset ($\rho=100$) using UTLO. The conditional images generated from the head (middle columns) and tail (right columns) classes share and are built on top of the same low-resolution (unconditional) images (left columns). Low-resolution images ($8 \times 8$) are upsampled to that of CIFAR10-LT ($32 \times 32$) for improved visualization.
   }
   \label{fig:cf10_uc}
    \vskip -0.15 in
\end{figure}

\noindent \textbf{Knowledge-sharing at Low Resolutions}.
As explained in Sec. \ref{sec:method}, our proposed generator builds on top of low-resolution (e.g., $8 \times 8$) unconditionally trained images which are subsequently used to generate conditional images at higher resolutions (e.g., $32 \times 32$). Several low-resolution images are shown in Fig. \ref{fig:cf10_uc} along with their high-resolution conditional images generated for the head and tail classes of the dataset. It can be seen that conditional images generated from the tail classes evidently share certain features from the head classes. This demonstrates that knowledge sharing at low resolutions is an effective approach for infusing information from head classes to tail ones. A quantitative analysis of knowledge-sharing is presented in Supp.~\ref{supp:knowledge_sharing}. For additional qualitative examples see Supp.~\ref{supp:H}. 

\textbf{CIFAR100-LT} In addition to CIFAR10-LT, we evaluated our method and various baselines on the more challenging CIFAR100-LT dataset, where the tail classes include as few as five training instances.
The quantitative results, shown in Table. \ref{table:cifar100}, reveal that our method outperforms the baselines in a benchmark where a large number of classes with high diversity are present. In Fig. \ref{fig:overfit}, we demonstrate the FID-FS curve during the course of the training along with the generated samples from the tail classes of the CIFAR100-LT dataset. This shows the effectiveness of our proposed method in addressing mode collapse while regularization methods diverge and require early stopping. Further visual comparisons of our proposed method against the baselines for CIFAR100-LT can be found in Supp.~\ref{supp:H}.

\begin{table}[h!]
\caption{Comparison of our proposed UTLO method with StyleGAN2-ADA baselines on the CIFAR100-LT dataset.}
\vskip -0.25in

\label{table:cifar100}
\begin{center}
\resizebox{1\linewidth}{!}{%

\begin{tabular}{l|cccc}

\multirow{2}{*}{Methods} & 
\multirow{2}{*}{FID $\downarrow$} &
\multirow{2}{*}{FID-FS $\downarrow$}  &
{KID $\downarrow$} & 
{KID-FS  $\downarrow$} \\

{} & \multicolumn{2}{c}{} & \multicolumn{2}{c}{$\times 1000$} \\

\hline \hline

{StyleGAN2-ADA \cite{karras2020training}}
&  {$10.8$}&{$24.9$} & {$5.1$}&{$9.3$} \\

{+ Transitional~\cite{shahbazi2022collapse}}
&  {${10.6}$}&{$23.7$} & {$\bb{4.2}$}&{$8.5$} \\

{+ GSR\cite{rangwani2022improving}}              
&  {${11.1}$}&{$25.0$} & {${5.0}$}&{$8.2$} \\

{+ NoisyTwins \cite{rangwani2023noisytwins}}
&  {$10.1$}&{$22.5$} & {$5.0$}&{$7.9$} \\

{+ UTLO (Ours)}                    
&  {$\bb{9.9}$}&{$\bb{21.8}$} & {${4.6}$}&{$\bb{7.5}$} \\

\end{tabular}
}
\end{center}
\vskip -0.35in
\end{table}

\textbf{LSUN5-LT} We include LSUN5-LT as a higher resolution ($128 \times 128$) and highly-imbalanced ($\rho=1000$) benchmark. We use Projected GAN with FastGAN\cite{liu2019large} generator and DA \cite{sauer2021projected}) as the baseline model. This is a different generator and discriminator design compared to StyleGAN2 along with a different augmentation method. In the conditional implementation of FastGAN, class-specific information is injected using class-conditional batch normalization, different from the style mapping network utilized in StyleGAN2. UTLO can be readily adapted to this design by using standard batch normalization layers at low resolutions and utilizing class-conditional batch normalization only at higher resolutions. However, it should be noted that methods such as \cite{rangwani2023noisytwins} are restricted to StyleGANs and cannot be applied in this context.

Table. \ref{table:lsun} compares the results obtained from our method against the baseline and with the addition of GSR \cite{rangwani2022improving}. Our method consistently surpasses the baselines by around $30\%$ across all metrics. As previously shown for CIFAR10-LT (Table \ref{table:cifar10}), UTLO not only improved on the tail classes but also significantly improved the dataset-wide FID/KID due to the high imbalance ratio ($\rho = 1000$).

Qualitative results in Fig. \ref{fig:lsun} contrast the generated images from our method against the baselines across all classes. For the \texttt{kitchen} class which only includes $50$ training examples, the baselines struggle to effectively learn from the tail class whereas UTLO succeeds in generating diverse and high-quality images. More visual examples from our method and baselines are presented in Supp.~\ref{supp:H}.

Due to space constraints, additional results and analysis are provided in the supplementary material. e.g., Analysis of the role of weighted sampling (Supp.~\ref{supp:ws}.), The distinction of training class-conditional GANs in the Long-tailed setup vs. Limited-data setup (Supp.~\ref{supp:E}.), and additional evaluation on Flowers-LT and iNaturalist2019 datasets (Supp.~\ref{supp:G}).

\begin{table}[t]
\caption{\small Evaluating the quality of generated images from the proposed method and comparing against baselines on LSUN5-LT.  }
\label{table:lsun}
\begin{center}
\vskip -0.2in
\begin{small}
\resizebox{1\linewidth}{!}
{

\begin{tabular}{l|cc|cc}

\multirow{2}{*}{Methods} & 
\multirow{2}{*}{FID $\downarrow$} &
\multirow{2}{*}{FID-FS $\downarrow$}  &
{KID $\downarrow$} & 
{KID-FS  $\downarrow$} \\ 

{} & \multicolumn{2}{c|}{} & \multicolumn{2}{c}{$\times 1000$}\\

\hline \hline
{PGAN (FastGAN)+DA~\cite{sauer2021projected}}            
&  {$15.0$}&{$60.2$}&{$4.6$}&{$52.7$} \\  

{+ GSR~\cite{rangwani2022improving}}              
&  {$15.7$}&{$63.7$}&{$5.7$}&{$58.0$} \\ 

{+ UTLO (Ours)}                    
& {$\bb{10.9}$}&{$\bb{43.6}$}&{$\bb{3.5}$}&{$\bb{35.3}$} \\ 

\end{tabular}
}
\end{small}
\end{center}
\vskip -0.35in
\end{table}

\vspace{-0.1in}
\section{Conclusion}
\vspace{-0.05in}
In this paper, we proposed UTLO, a novel framework designed to improve the training of cGANs on long-tailed data. Inspired by the observation that head and tail classes often have more similarities at lower resolutions, our method facilitates knowledge sharing from head to tail classes using unconditional training at lower resolutions. The proposed method enriches the limited training distribution of the tail classes and effectively addresses mode collapse, leading to significant improvement in image generation for tail classes. We have also introduced the FID-FS/KID-FS metric, an adaptation of widely-used GAN metrics, specifically tailored for tail classes. We hope that our findings are useful for future work in long-tail learning.

\vspace{-0.1in}
\section*{Acknowledgements}
This work is partially supported by the National Science Foundation (NSF) under grant 1751412 and 1927564.

{
    \small
    \bibliographystyle{ieeenat_fullname}
    \bibliography{main}

\begin{thebibliography}{59}
\providecommand{\natexlab}[1]{#1}
\providecommand{\url}[1]{\texttt{#1}}
\expandafter\ifx\csname urlstyle\endcsname\relax
  \providecommand{\doi}[1]{doi: #1}\else
  \providecommand{\doi}{doi: \begingroup \urlstyle{rm}\Url}\fi

\bibitem[Bi{\'n}kowski et~al.(2018)Bi{\'n}kowski, Sutherland, Arbel, and Gretton]{binkowski2018demystifying}
Miko{\l}aj Bi{\'n}kowski, Danica~J Sutherland, Michael Arbel, and Arthur Gretton.
\newblock Demystifying mmd gans.
\newblock In \emph{International Conference on Learning Representations}, 2018.

\bibitem[Brock et~al.(2019)Brock, Donahue, and Simonyan]{DBLP:conf/iclr/BrockDS19}
Andrew Brock, Jeff Donahue, and Karen Simonyan.
\newblock Large scale {GAN} training for high fidelity natural image synthesis.
\newblock In \emph{7th International Conference on Learning Representations, {ICLR} 2019, New Orleans, LA, USA, May 6-9, 2019}. OpenReview.net, 2019.

\bibitem[Cao et~al.(2019)Cao, Wei, Gaidon, Arechiga, and Ma]{NEURIPS2019_621461af}
Kaidi Cao, Colin Wei, Adrien Gaidon, Nikos Arechiga, and Tengyu Ma.
\newblock Learning imbalanced datasets with label-distribution-aware margin loss.
\newblock In \emph{Advances in Neural Information Processing Systems}. Curran Associates, Inc., 2019.

\bibitem[Cui et~al.(2021)Cui, Liu, Tian, Zhong, and Jia]{cui2021reslt}
Jiequan Cui, Shu Liu, Zhuotao Tian, Zhisheng Zhong, and Jiaya Jia.
\newblock Reslt: Residual learning for long-tailed recognition.
\newblock \emph{ieee transactions on pattern analysis and machine intelligence}, 2021.

\bibitem[Cui et~al.(2019)Cui, Jia, Lin, Song, and Belongie]{DBLP:conf/cvpr/CuiJLSB19}
Yin Cui, Menglin Jia, Tsung{-}Yi Lin, Yang Song, and Serge~J. Belongie.
\newblock Class-balanced loss based on effective number of samples.
\newblock In \emph{{IEEE} Conference on Computer Vision and Pattern Recognition, {CVPR} 2019, Long Beach, CA, USA, June 16-20, 2019}, pages 9268--9277. Computer Vision Foundation / {IEEE}, 2019.

\bibitem[Dhariwal and Nichol(2021)]{DBLP:conf/nips/DhariwalN21}
Prafulla Dhariwal and Alexander~Quinn Nichol.
\newblock Diffusion models beat gans on image synthesis.
\newblock In \emph{Advances in Neural Information Processing Systems 34: Annual Conference on Neural Information Processing Systems 2021, NeurIPS 2021, December 6-14, 2021, virtual}, pages 8780--8794, 2021.

\bibitem[Goodfellow et~al.(2014)Goodfellow, Pouget-Abadie, Mirza, Xu, Warde-Farley, Ozair, Courville, and Bengio]{goodfellowGAN}
Ian Goodfellow, Jean Pouget-Abadie, Mehdi Mirza, Bing Xu, David Warde-Farley, Sherjil Ozair, A~aron Courville, and Yoshua Bengio.
\newblock Generative adversarial nets.
\newblock In \emph{Advances in Neural Information Processing Systems}. Curran Associates, Inc., 2014.

\bibitem[Guo and Wang(2021)]{guo2021long}
Hao Guo and Song Wang.
\newblock Long-tailed multi-label visual recognition by collaborative training on uniform and re-balanced samplings.
\newblock In \emph{Proceedings of the IEEE/CVF Conference on Computer Vision and Pattern Recognition}, pages 15089--15098, 2021.

\bibitem[Han et~al.(2005)Han, Wang, and Mao]{han2005borderline}
Hui Han, Wen-Yuan Wang, and Bing-Huan Mao.
\newblock Borderline-smote: a new over-sampling method in imbalanced data sets learning.
\newblock In \emph{International conference on intelligent computing}, pages 878--887. Springer, 2005.

\bibitem[Heusel et~al.(2017)Heusel, Ramsauer, Unterthiner, Nessler, and Hochreiter]{heusel2017gans}
Martin Heusel, Hubert Ramsauer, Thomas Unterthiner, Bernhard Nessler, and Sepp Hochreiter.
\newblock Gans trained by a two time-scale update rule converge to a local nash equilibrium.
\newblock \emph{Advances in neural information processing systems}, 30, 2017.

\bibitem[Horn and Aodha()]{iNaturalist2019}
Grant~Van Horn and Oisin~Mac Aodha.
\newblock The inaturalist 2019 competition dataset.
\newblock \url{https://www.kaggle.com/c/inaturalist-2019-fgvc6}.

\bibitem[Isola et~al.(2017)Isola, Zhu, Zhou, and Efros]{isola2017image}
Phillip Isola, Jun-Yan Zhu, Tinghui Zhou, and Alexei~A Efros.
\newblock Image-to-image translation with conditional adversarial networks.
\newblock In \emph{Proceedings of the IEEE conference on computer vision and pattern recognition}, pages 1125--1134, 2017.

\bibitem[Kang et~al.(2020)Kang, Xie, Rohrbach, Yan, Gordo, Feng, and Kalantidis]{DBLP:conf/iclr/KangXRYGFK20}
Bingyi Kang, Saining Xie, Marcus Rohrbach, Zhicheng Yan, Albert Gordo, Jiashi Feng, and Yannis Kalantidis.
\newblock Decoupling representation and classifier for long-tailed recognition.
\newblock In \emph{8th International Conference on Learning Representations, {ICLR} 2020, Addis Ababa, Ethiopia, April 26-30, 2020}. OpenReview.net, 2020.

\bibitem[Kang et~al.(2021)Kang, Li, Xie, Yuan, and Feng]{Kang2021ExploringBF}
Bingyi Kang, Yu Li, Sai~Nan Xie, Zehuan Yuan, and Jiashi Feng.
\newblock Exploring balanced feature spaces for representation learning.
\newblock In \emph{International Conference on Learning Representations}, 2021.

\bibitem[Karmali et~al.(2022)Karmali, Parihar, Agrawal, Rangwani, Jampani, Singh, and Babu]{KarmaliPARJ0B22}
Tejan Karmali, Rishubh Parihar, Susmit Agrawal, Harsh Rangwani, Varun Jampani, Maneesh~Kumar Singh, and R.~Venkatesh Babu.
\newblock Hierarchical semantic regularization of latent spaces in stylegans.
\newblock In \emph{Computer Vision - {ECCV} 2022 - 17th European Conference, Tel Aviv, Israel, October 23-27, 2022, Proceedings, Part {XV}}, pages 443--459. Springer, 2022.

\bibitem[Karnewar and Wang(2020)]{DBLP:conf/cvpr/KarnewarW20}
Animesh Karnewar and Oliver Wang.
\newblock {MSG-GAN:} multi-scale gradients for generative adversarial networks.
\newblock In \emph{2020 {IEEE/CVF} Conference on Computer Vision and Pattern Recognition, {CVPR} 2020, Seattle, WA, USA, June 13-19, 2020}, pages 7796--7805. Computer Vision Foundation / {IEEE}, 2020.

\bibitem[Karras et~al.(2020{\natexlab{a}})Karras, Aittala, Hellsten, Laine, Lehtinen, and Aila]{karras2020training}
Tero Karras, Miika Aittala, Janne Hellsten, Samuli Laine, Jaakko Lehtinen, and Timo Aila.
\newblock Training generative adversarial networks with limited data.
\newblock \emph{Advances in Neural Information Processing Systems}, 33:\penalty0 12104--12114, 2020{\natexlab{a}}.

\bibitem[Karras et~al.(2020{\natexlab{b}})Karras, Laine, Aittala, Hellsten, Lehtinen, and Aila]{stylegan2Karras}
Tero Karras, Samuli Laine, Miika Aittala, Janne Hellsten, Jaakko Lehtinen, and Timo Aila.
\newblock Analyzing and improving the image quality of stylegan.
\newblock In \emph{2020 {IEEE/CVF} Conference on Computer Vision and Pattern Recognition, {CVPR} 2020, Seattle, WA, USA, June 13-19, 2020}, pages 8107--8116. Computer Vision Foundation / {IEEE}, 2020{\natexlab{b}}.

\bibitem[Kong et~al.(2022)Kong, Kim, Han, and Kwak]{KongKHK22smoothing}
Chaerin Kong, Jeesoo Kim, Donghoon Han, and Nojun Kwak.
\newblock Few-shot image generation with mixup-based distance learning.
\newblock In \emph{Computer Vision - {ECCV} 2022 - 17th European Conference, Tel Aviv, Israel, October 23-27, 2022, Proceedings, Part {XV}}, pages 563--580. Springer, 2022.

\bibitem[Krizhevsky et~al.(2009)Krizhevsky, Hinton, et~al.]{krizhevsky2009learning}
Alex Krizhevsky, Geoffrey Hinton, et~al.
\newblock Learning multiple layers of features from tiny images.
\newblock \emph{preprint}, 2009.

\bibitem[Kumari et~al.(2022)Kumari, Zhang, Shechtman, and Zhu]{kumari2022ensembling}
Nupur Kumari, Richard Zhang, Eli Shechtman, and Jun-Yan Zhu.
\newblock Ensembling off-the-shelf models for gan training.
\newblock In \emph{Proceedings of the IEEE/CVF Conference on Computer Vision and Pattern Recognition}, pages 10651--10662, 2022.

\bibitem[Ledig et~al.(2017)Ledig, Theis, Husz{\'a}r, Caballero, Cunningham, Acosta, Aitken, Tejani, Totz, Wang, et~al.]{ledig2017photo}
Christian Ledig, Lucas Theis, Ferenc Husz{\'a}r, Jose Caballero, Andrew Cunningham, Alejandro Acosta, Andrew Aitken, Alykhan Tejani, Johannes Totz, Zehan Wang, et~al.
\newblock Photo-realistic single image super-resolution using a generative adversarial network.
\newblock In \emph{Proceedings of the IEEE conference on computer vision and pattern recognition}, pages 4681--4690, 2017.

\bibitem[Liu et~al.(2021)Liu, Zhu, Song, and Elgammal]{liu2021towards}
Bingchen Liu, Yizhe Zhu, Kunpeng Song, and A. Elgammal.
\newblock Towards faster and stabilized gan training for high-fidelity few-shot image synthesis.
\newblock \emph{iclr}, 2021.

\bibitem[Liu et~al.(2020)Liu, Sun, Han, Dou, and Li]{liu2020deep}
Jialun Liu, Yifan Sun, Chuchu Han, Zhaopeng Dou, and Wenhui Li.
\newblock Deep representation learning on long-tailed data: A learnable embedding augmentation perspective.
\newblock \emph{Computer Vision And Pattern Recognition}, 2020.

\bibitem[Liu and Zhang(2020)]{liu2020under}
Shudong Liu and Ke Zhang.
\newblock Under-sampling and feature selection algorithms for s2smlp.
\newblock \emph{IEEE Access}, 8:\penalty0 191803--191814, 2020.

\bibitem[Liu et~al.(2019)Liu, Miao, Zhan, Wang, Gong, and Yu]{liu2019large}
Ziwei Liu, Zhongqi Miao, Xiaohang Zhan, Jiayun Wang, Boqing Gong, and Stella~X Yu.
\newblock Large-scale long-tailed recognition in an open world.
\newblock In \emph{Proceedings of the IEEE/CVF Conference on Computer Vision and Pattern Recognition}, pages 2537--2546, 2019.

\bibitem[Lucic et~al.(2018)Lucic, Kurach, Michalski, Gelly, and Bousquet]{lucic2018gans}
Mario Lucic, Karol Kurach, Marcin Michalski, Sylvain Gelly, and Olivier Bousquet.
\newblock Are gans created equal? a large-scale study.
\newblock \emph{Advances in neural information processing systems}, 31, 2018.

\bibitem[Menon et~al.(2021)Menon, Jayasumana, Rawat, Jain, Veit, and Kumar]{menon2021longtail}
Aditya~Krishna Menon, Sadeep Jayasumana, Ankit~Singh Rawat, Himanshu Jain, Andreas Veit, and Sanjiv Kumar.
\newblock Long-tail learning via logit adjustment.
\newblock In \emph{International Conference on Learning Representations}, 2021.

\bibitem[Mirza and Osindero(2014)]{mirza2014conditional}
Mehdi Mirza and Simon Osindero.
\newblock Conditional generative adversarial nets.
\newblock \emph{arXiv preprint arXiv: Arxiv-1411.1784}, 2014.

\bibitem[Miyato et~al.(2018)Miyato, Kataoka, Koyama, and Yoshida]{miyato2018spectral}
Takeru Miyato, Toshiki Kataoka, Masanori Koyama, and Yuichi Yoshida.
\newblock Spectral normalization for generative adversarial networks.
\newblock \emph{ICLR}, 2018.

\bibitem[Mullick et~al.(2019)Mullick, Datta, and Das]{mullick2019generative}
S.~S. Mullick, Shounak Datta, and Swagatam Das.
\newblock Generative adversarial minority oversampling.
\newblock \emph{Ieee International Conference On Computer Vision}, 2019.

\bibitem[Nilsback and Zisserman(2008)]{Nilsback08}
Maria-Elena Nilsback and Andrew Zisserman.
\newblock Automated flower classification over a large number of classes.
\newblock In \emph{Indian Conference on Computer Vision, Graphics and Image Processing}, 2008.

\bibitem[Odena et~al.(2017)Odena, Olah, and Shlens]{OdenaOS17ACGAN}
Augustus Odena, Christopher Olah, and Jonathon Shlens.
\newblock Conditional image synthesis with auxiliary classifier gans.
\newblock In \emph{Proceedings of the 34th International Conference on Machine Learning, {ICML} 2017, Sydney, NSW, Australia, 6-11 August 2017}, pages 2642--2651. {PMLR}, 2017.

\bibitem[Ojha et~al.(2021)Ojha, Li, Lu, Efros, Lee, Shechtman, and Zhang]{Ojha_2021_CVPR}
Utkarsh Ojha, Yijun Li, Jingwan Lu, Alexei~A. Efros, Yong~Jae Lee, Eli Shechtman, and Richard Zhang.
\newblock Few-shot image generation via cross-domain correspondence.
\newblock In \emph{Proceedings of the IEEE/CVF Conference on Computer Vision and Pattern Recognition (CVPR)}, pages 10743--10752, 2021.

\bibitem[Rangwani et~al.(2021)Rangwani, Mopuri, and Babu]{pmlr-v161-rangwani21a}
Harsh Rangwani, Konda~Reddy Mopuri, and R.~Venkatesh Babu.
\newblock Class balancing gan with a classifier in the loop.
\newblock In \emph{Proceedings of the Thirty-Seventh Conference on Uncertainty in Artificial Intelligence}, pages 1618--1627. PMLR, 2021.

\bibitem[Rangwani et~al.(2022)Rangwani, Jaswani, Karmali, Jampani, and Babu]{rangwani2022improving}
Harsh Rangwani, Naman Jaswani, Tejan Karmali, Varun Jampani, and R.~Venkatesh Babu.
\newblock Improving gans for long-tailed data through group spectral regularization.
\newblock \emph{European Conference On Computer Vision}, 2022.

\bibitem[Rangwani et~al.(2023)Rangwani, Bansal, Sharma, Karmali, Jampani, and Babu]{rangwani2023noisytwins}
Harsh Rangwani, Lavish Bansal, Kartik Sharma, Tejan Karmali, Varun Jampani, and R~Venkatesh Babu.
\newblock Noisytwins: Class-consistent and diverse image generation through stylegans.
\newblock In \emph{Proceedings of the IEEE/CVF Conference on Computer Vision and Pattern Recognition}, pages 5987--5996, 2023.

\bibitem[Ren et~al.(2020)Ren, Yu, sheng, Ma, Zhao, Yi, and Li]{NEURIPS2020_2ba61cc3}
Jiawei Ren, Cunjun Yu, shunan sheng, Xiao Ma, Haiyu Zhao, Shuai Yi, and hongsheng Li.
\newblock Balanced meta-softmax for long-tailed visual recognition.
\newblock In \emph{Advances in Neural Information Processing Systems}, pages 4175--4186. Curran Associates, Inc., 2020.

\bibitem[Santurkar et~al.(2017)Santurkar, Schmidt, and Madry]{santurkar2017classificationbased}
Shibani Santurkar, Ludwig Schmidt, and A. Madry.
\newblock A classification-based study of covariate shift in gan distributions.
\newblock \emph{International Conference On Machine Learning}, 2017.

\bibitem[Sauer et~al.(2021)Sauer, Chitta, M{\"u}ller, and Geiger]{sauer2021projected}
Axel Sauer, Kashyap Chitta, Jens M{\"u}ller, and Andreas Geiger.
\newblock Projected gans converge faster.
\newblock \emph{Advances in Neural Information Processing Systems}, 34:\penalty0 17480--17492, 2021.

\bibitem[Sauer et~al.(2022)Sauer, Schwarz, and Geiger]{SauerS022StyleGAN-XL}
Axel Sauer, Katja Schwarz, and Andreas Geiger.
\newblock Stylegan-xl: Scaling stylegan to large diverse datasets.
\newblock In \emph{{SIGGRAPH} '22: Special Interest Group on Computer Graphics and Interactive Techniques Conference, Vancouver, BC, Canada, August 7 - 11, 2022}, pages 49:1--49:10. {ACM}, 2022.

\bibitem[Shahbazi et~al.(2022)Shahbazi, Danelljan, Paudel, and Gool]{shahbazi2022collapse}
Mohamad Shahbazi, Martin Danelljan, Danda~Pani Paudel, and Luc~Van Gool.
\newblock Collapse by conditioning: Training class-conditional {GAN}s with limited data.
\newblock In \emph{International Conference on Learning Representations}, 2022.

\bibitem[Shen and Zhou(2021)]{shen2021closedform}
Yujun Shen and Bolei Zhou.
\newblock Closed-form factorization of latent semantics in gans.
\newblock In \emph{CVPR}, 2021.

\bibitem[Si and Zhu(2012)]{animalfaces}
Zhangzhang Si and Song-Chun Zhu.
\newblock Learning hybrid image templates (hit) by information projection.
\newblock \emph{IEEE Transactions on Pattern Analysis and Machine Intelligence}, 34\penalty0 (7):\penalty0 1354--1367, 2012.

\bibitem[Tang et~al.(2020)Tang, Huang, and Zhang]{NEURIPS2020_1091660f}
Kaihua Tang, Jianqiang Huang, and Hanwang Zhang.
\newblock Long-tailed classification by keeping the good and removing the bad momentum causal effect.
\newblock In \emph{Advances in Neural Information Processing Systems}, pages 1513--1524. Curran Associates, Inc., 2020.

\bibitem[Tran et~al.(2021)Tran, Tran, Nguyen, Nguyen, and Cheung]{tran2021data}
Ngoc-Trung Tran, Viet-Hung Tran, Ngoc-Bao Nguyen, Trung-Kien Nguyen, and Ngai-Man Cheung.
\newblock On data augmentation for gan training.
\newblock \emph{IEEE Transactions on Image Processing}, 30:\penalty0 1882--1897, 2021.

\bibitem[Tseng et~al.(2021)Tseng, Jiang, Liu, Yang, and Yang]{tseng2021regularizing}
Hung-Yu Tseng, Lu Jiang, Ce Liu, Ming-Hsuan Yang, and Weilong Yang.
\newblock Regularizing generative adversarial networks under limited data.
\newblock In \emph{Proceedings of the IEEE/CVF Conference on Computer Vision and Pattern Recognition}, pages 7921--7931, 2021.

\bibitem[Wang et~al.(2021)Wang, Lian, Miao, Liu, and Yu]{DBLP:conf/iclr/WangLM0Y21}
Xudong Wang, Long Lian, Zhongqi Miao, Ziwei Liu, and Stella~X. Yu.
\newblock Long-tailed recognition by routing diverse distribution-aware experts.
\newblock In \emph{9th International Conference on Learning Representations, {ICLR} 2021, Virtual Event, Austria, May 3-7, 2021}. OpenReview.net, 2021.

\bibitem[Yang et~al.(2022)Yang, Jiang, Song, and Guo]{yang22surveyLT}
Lu Yang, He Jiang, Qing Song, and Jun Guo.
\newblock A survey on long-tailed visual recognition.
\newblock \emph{International Journal of Computer Vision}, 130:\penalty0 1837--1872, 2022.

\bibitem[Yao and Wang(2021)]{yao2021improved}
Baofeng Yao and Lei Wang.
\newblock An improved under-sampling imbalanced classification algorithm.
\newblock In \emph{2021 13th International Conference on Measuring Technology and Mechatronics Automation (ICMTMA)}, pages 775--779. IEEE, 2021.

\bibitem[Yu et~al.(2015)Yu, Zhang, Song, Seff, and Xiao]{yu15lsun}
Fisher Yu, Yinda Zhang, Shuran Song, Ari Seff, and Jianxiong Xiao.
\newblock Lsun: Construction of a large-scale image dataset using deep learning with humans in the loop.
\newblock \emph{arXiv preprint arXiv:1506.03365}, 2015.

\bibitem[Zhang et~al.(2017)Zhang, Xu, Li, Zhang, Wang, Huang, and Metaxas]{zhang2017stackgan}
Han Zhang, Tao Xu, Hongsheng Li, Shaoting Zhang, Xiaogang Wang, Xiaolei Huang, and Dimitris~N Metaxas.
\newblock Stackgan: Text to photo-realistic image synthesis with stacked generative adversarial networks.
\newblock In \emph{Proceedings of the IEEE international conference on computer vision}, pages 5907--5915, 2017.

\bibitem[Zhang et~al.(2019)Zhang, Zhang, Odena, and Lee]{zhang2019consistency}
Han Zhang, Zizhao Zhang, Augustus Odena, and Honglak Lee.
\newblock Consistency regularization for generative adversarial networks.
\newblock In \emph{International Conference on Learning Representations}, 2019.

\bibitem[Zhang et~al.(2018)Zhang, Isola, Efros, Shechtman, and Wang]{zhang2018unreasonable}
Richard Zhang, Phillip Isola, Alexei~A. Efros, Eli Shechtman, and Oliver Wang.
\newblock The unreasonable effectiveness of deep features as a perceptual metric.
\newblock \emph{Ieee/cvf Conference On Computer Vision And Pattern Recognition}, 2018.

\bibitem[Zhang et~al.(2021)Zhang, Kang, Hooi, Yan, and Feng]{zhang2021deepLT}
Yifan Zhang, Bingyi Kang, Bryan Hooi, Shuicheng Yan, and Jiashi Feng.
\newblock Deep long-tailed learning: A survey.
\newblock \emph{arXiv preprint arXiv: Arxiv-2110.04596}, 2021.

\bibitem[Zhang and Pfister(2021)]{zhang2021learning}
Zizhao Zhang and Tomas Pfister.
\newblock Learning fast sample re-weighting without reward data.
\newblock In \emph{Proceedings of the IEEE/CVF International Conference on Computer Vision}, pages 725--734, 2021.

\bibitem[Zhao et~al.(2020{\natexlab{a}})Zhao, Liu, Lin, Zhu, and Han]{zhao2020differentiable}
Shengyu Zhao, Zhijian Liu, Ji Lin, Jun-Yan Zhu, and Song Han.
\newblock Differentiable augmentation for data-efficient gan training.
\newblock \emph{Advances in Neural Information Processing Systems}, 33:\penalty0 7559--7570, 2020{\natexlab{a}}.

\bibitem[Zhao et~al.(2020{\natexlab{b}})Zhao, Zhang, Chen, Singh, and Zhang]{zhao2020image}
Zhengli Zhao, Zizhao Zhang, Ting Chen, Sameer Singh, and Han Zhang.
\newblock Image augmentations for gan training.
\newblock \emph{arXiv preprint arXiv: 2006.02595}, 2020{\natexlab{b}}.

\bibitem[Zhong et~al.(2021)Zhong, Cui, Liu, and Jia]{DBLP:conf/cvpr/ZhongC0J21}
Zhisheng Zhong, Jiequan Cui, Shu Liu, and Jiaya Jia.
\newblock Improving calibration for long-tailed recognition.
\newblock In \emph{{IEEE} Conference on Computer Vision and Pattern Recognition, {CVPR} 2021, virtual, June 19-25, 2021}, pages 16489--16498. Computer Vision Foundation / {IEEE}, 2021.

\end{thebibliography}
}

\appendix
\onecolumn

\section*{\center \huge Supplementary Materials}

\section{Datasets}\label{supp:dataset}
We use 6 diverse long-tailed datasets in our experiments: CIFAR10 and CIFAR100\cite{krizhevsky2009learning}, LSUN \cite{yu15lsun}, Flowers \cite{Nilsback08}, iNaturalist2019 \cite{iNaturalist2019}, and AnimalFaces \cite{animalfaces} to encompass a wide range of image domains, dataset sizes, resolutions, and imbalanced ratios $\rho$. In the following, we provide a detailed description of each dataset.

\begin{itemize}
    \item \textbf{CIFAR10/100:} The original CIFAR10 and CIFAR100 datasets consist of 50,000 training images at $32 \times 32$ resolution with an equal number of images present in 10 and 100 classes, respectively. An imbalanced version of these datasets referred to as CIFAR10/100-LT, is widely used in the long-tail recognition literature \cite{yang22surveyLT}. We follow \cite{DBLP:conf/cvpr/CuiJLSB19} to generate an exponentially long-tailed version with an imbalance ratio  $\rho=\{50,100\}$ for CIFAR10-LT and $\rho=100$ for CIFAR100-LT. When calculating few-shot metrics, we follow \cite{cui2021reslt} and take classes 6-9 in CIFAR10-LT and 70-99 in CIFAR100-LT as the few-shot subset in our evaluation.
    
    \item \textbf{LSUN}:
    Following \cite{santurkar2017classificationbased, rangwani2022improving}, we select a challenging subset of 5 classes in the LSUN Scene dataset and keep 50,000 images from the training set (250,000 in total): \texttt{dining room}, \texttt{conference room}, \texttt{bedroom}, \texttt{living room}, and \texttt{kitchen}. We make this subset long-tailed with an imbalance ratio $\rho=1000$ and refer to it as LSUN5-LT. We take the \texttt{kitchen} class with only 50 training samples as the few-shot subset.
    
    \item \textbf{Flowers}: Oxford Flowers dataset contains 102 different flower categories. We first combine the train and validation set images to access a larger dataset for the purpose of training and evaluation. This results in a total of 7370 images across all classes. This dataset is naturally imbalanced with 234 and 34 images being present in the category with the most and least number of images ($\rho \approx 7$), respectively. To increase the skewness and learning difficulty, we further increase the imbalance ratio to $\rho = 100$. This reduces the number of training images in the tail classes to only 2 images. We use $128 \times 128$ resolution in the experiments and refer to this as Flowers-LT. For the few-shot reference, we take the 52 classes with the least number of training samples (classes 51-102), containing from 23 to 2 images.
    
    \item \textbf{iNaturalist2019}: The 2019 version of the iNaturalist dataset is a large-scale fine-grained dataset containing 1,010 species from nature. This dataset naturally follows a long-tailed distribution and we keep the training instances in each class intact. We use the training set at the $64 \times 64$ resolution in the experiments. The training set of iNaturalist2019 contains a total of 268,243 images. We take the 210 classes with the fewest training instances for the few-shot evaluation (classes 801-1010).
 
    \item \textbf{AnimalFaces}: This dataset contains the faces of 20 different animal categories (including humans). We set the resolution of the images to $64 \times 64$. The most populated categories in the dataset are \texttt{dog} and \texttt{cat} classes with 388 and 160 images, respectively. The remaining 18 classes are roughly balanced containing from 118 to 100 images. This indicates an imbalance ratio close to 4 ($\rho\approx4$). To make it more suitable for the long-tail setup, we artificially increase the imbalance ratio to $\rho=25$, increasing the training difficulty. We refer to this as AnimalFaces-LT in our experiments. For few-shot evaluation, we take the 10 least frequent categories (classes 11-20).

\end{itemize}

Across all datasets, we sort image categories in the decreasing order of their size, i.e., class index 0 containing the most training images and so on.

\section{Effects of Weighted Sampling}\label{supp:ws}
\begin{figure}[bht]
  \centering
  \includegraphics[width=0.5\linewidth]{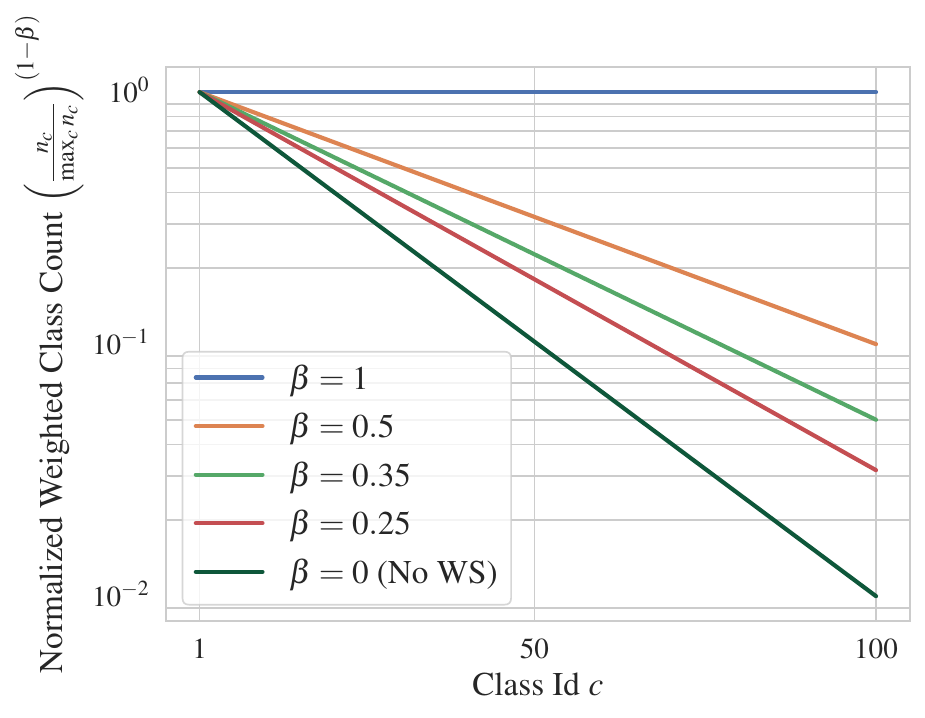}
  \caption{WS CIFAR100-LT $(\rho=100)$. When $\beta=0$, the sampling follows the original data distribution. As $\beta$ increases, the \textit{effective} imbalance ratio $\varrho$ among the classes decreases. At $\beta=1$, head and tail classes have the same probability of being drawn during training, i.e. $\varrho = 1$. }
  \label{fig:ws}
\end{figure}

   Long-tailed datasets suffer from severe class imbalance problems that hinder learning, especially for the tail classes. To mitigate this issue, one common technique in the long-tail recognition literature \cite{yang22surveyLT} is to adjust the sampling during training by assigning higher weights to the tail classes, i.e., oversampling them. With recent advancements in data augmentation for GAN training \cite{zhang2019consistency, karras2020training, zhao2020differentiable}, it is reasonable to speculate that oversampling tail instances might improve training calibration. To investigate this, we choose a simple yet effective weighted sampling (WS) method to compare against traditional random sampling methods by assigning a weight $w_c$ for the samples from class $c \in \{1, \dots, C\}$,
\begin{align}
    w_c = {n_c ^{-\beta}}, \quad \varrho = \rho^{(1 - \beta)}
\end{align} \label{eq:weight_sampler}
where $n_c$ is the number of samples in class $c$, $\rho = \max_c\{n_c\}/\min_c\{n_c\}$ is the imbalance ratio, and $\beta \in [0,1]$ is a hyperparameter that damps the sampling imbalance. We illustrate this in Fig. \ref{fig:ws} for CIFAR100-LT with $\rho=100$. When $\beta=0$, the sampling follows the original data distribution. As $\beta$ increases, the \textit{effective} imbalance ratio $\varrho$ among the classes decreases. At $\beta=1$, head and tail classes have the same probability of being drawn during training, i.e. $\varrho = 1$. While WS might help with more calibrated training, it does not promote knowledge sharing among classes.\\

\begin{figure}[bht]
  \centering
  \includegraphics[width=0.45\linewidth]{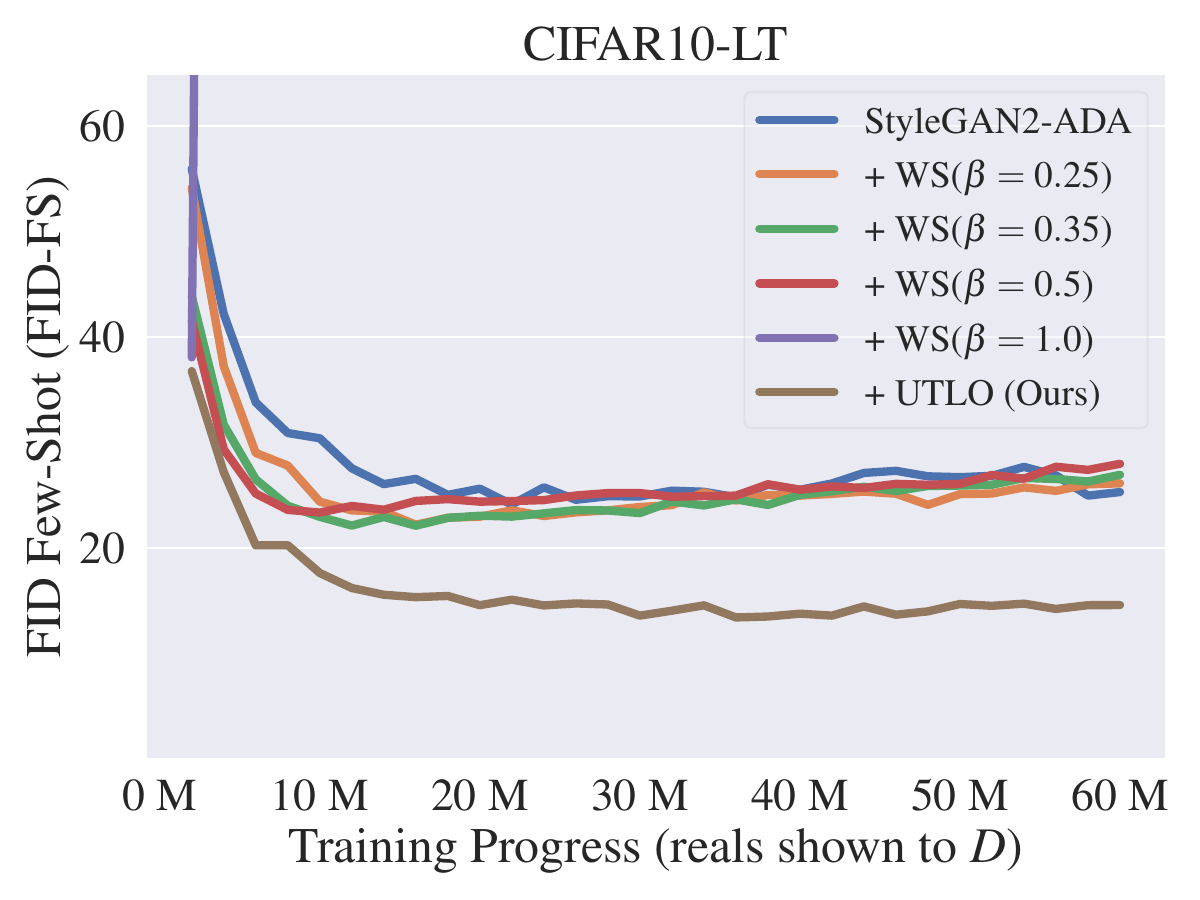}
  \includegraphics[width=0.45\linewidth]{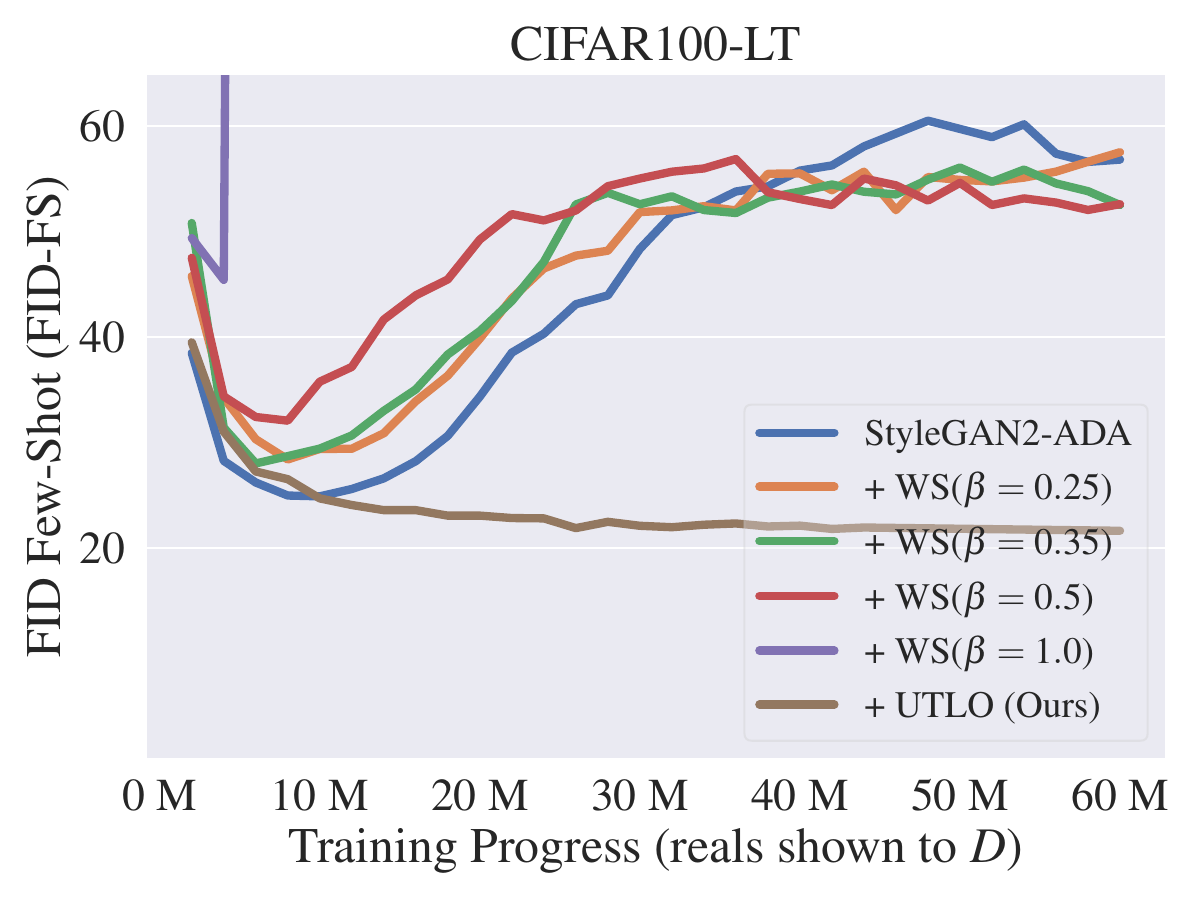}
  \caption{FID-FS curves for CIFAR10-LT and CIFAR100-LT datasets when using weighted sampling (WS) with different $\beta$ values. We observe that WS leads to overfitting, and as $\beta$ increases, this becomes more pronounced. Training becomes unstable when $\beta=1$ for both datasets.}
  \label{fig:cf_ws}
\end{figure}

Table. \ref{table:ws} presents a comparison of the results obtained from the StyleGAN2-ADA baseline and the use of weighted sampling (WS). Our experiments indicate that WS improves the FID and KID metrics compared to the baseline. However, it does not enhance tail performance in terms of FID-FS and KID-FS, indicating that balancing alone is insufficient when there are very limited training examples in the tail. We also noticed that adding weighted sampling will boost overfitting. Fig. \ref{fig:cf_ws} illustrates the training FID-FS curves for the WS methods with different $\beta$ values, compared against our method and the baseline. For both CIFAR10-LT ($\rho=100$) and CIFAR100-LT ($\rho=100$) datasets, WS methods exhibit overfitting behavior. As $\beta$ increases, this becomes more pronounced, and training becomes unstable when $\beta=1$ across both datasets. While WS did not show any improvements for CIFAR100-LT, we found it to demonstrate relatively better performance than the baseline at the early stages of training before overfitting sets in. While our experiments on the role of WS in training cGANs in the long-tail setup are informative, we believe that reaching a comprehensive conclusion requires further analysis.

\begin{table}[bht]

\caption{\small Effect of weighted-sampling (WS) when training on long-tailed datasets.}

\label{table:ws}
\begin{center}
\resizebox{0.9\textwidth}{!}{%

\begin{tabular}{l|cccc|cccc}

{Dataset} & \multicolumn{4}{c}{CIFAR10-LT} & \multicolumn{4}{c}{CIFAR100-LT}\\
\hline\hline

\multirow{2}{*}{Metrics} & 
\multirow{2}{*}{FID $\downarrow$} &
\multirow{2}{*}{FID-FS $\downarrow$}  &
{KID $\downarrow$} & 
{KID-FS  $\downarrow$} &
\multirow{2}{*}{FID $\downarrow$} &
\multirow{2}{*}{FID-FS $\downarrow$}  &
{KID $\downarrow$} & 
{KID-FS  $\downarrow$} \\

{} & \multicolumn{2}{c}{} & \multicolumn{2}{c|}{$\times 1000$} &  \multicolumn{2}{c}{} &  \multicolumn{2}{c}{$\times 1000$}\\

\hline \hline
{StyleGAN2-ADA \cite{karras2020training}}
&  {$9.0$}&{$24.2$} & {$4.0$}&{$9.7$} 
&  {$10.8$}&{$24.9$} & {$5.1$}&{$9.3$} \\

{+ GSR\cite{rangwani2022improving}}              
&  {${8.4}$}&{$24.3$} & {${3.9}$}&{$11.8$} 
&  {${11.1}$}&{$25.0$} & {${5.0}$}&{$8.2$} \\

{+ UTLO (Ours)}                    
&  {$\bb{6.8}$}&{$\bb{13.4}$} & {${2.8}$}&{$\bb{5.4}$}
&  {$\bb{9.9}$}&{$\bb{21.8}$} & {$\bb{4.6}$}&{$\bb{7.5}$} \\

\hline
{+ WS ($\beta = 0.25$)}
&  {$7.1$}&{$22.2$}&{$2.6$}&{$8.1$} 
&  {$13.7$}&{$28.4$}&{$7.2$}&{$10.5$} \\

{+ WS ($\beta = 0.35$)}
&  {$7.6$}&{$22.1$}&{$2.9$}&{$8.0$}
&  {$14.1$}&{$28.0$}&{$7.0$}&{$9.4$} \\

{+ WS ($\beta = 0.5$)}
&  {$8.0$}&{$23.4$}&{$\bb{2.5}$}&{$8.0$}
&  {$14.9$}&{$32.1$}&{$7.3$}&{$10.4$} \\

\end{tabular}
}
\end{center}
\vskip -0.2in
\end{table}

\section{Ablation Study}\label{supp:ablation}

\textbf{Choice of Intermediate Low Resolution.} In our proposed method, one of the hyperparameter choices is to select an intermediate low resolution $res_{\text{uc}}$ for unconditional training. All layers with equal or lower resolution than $res_{\text{uc}}$ do not have access to class-conditional information. An unconditional GAN objective is added over the images and/or features at $res_{\text{uc}}$. To study the impact of $res_{\text{uc}}$, we conducted an ablation study on the AnimalFaces-LT dataset, which contains images at $64 \times 64$. Table. \ref{table:ablation_res} presents the results for selecting $res_{\text{uc}}$ from resolutions lower than $64 \times 64$: $8 \times 8$, $16 \times 16$, and $32 \times 32$. For all resolutions, we set the unconditional objective weight $\lambda = 1$.\\

Studying the results obtained in Table. \ref{table:ablation_res}, we observe that resolutions of $8 \times 8$ and $16 \time 16$ achieve relatively close performance. On the other hand, the performance degrades when layers up to $32 \times 32$ are trained unconditionally. This is anticipated as the AnimalFaces-LT is at $64 \times 64$, leaving only one layer to learn the class-conditional information which is shown to be insufficient. \\

\begin{figure}[bht]
  \centering
  \includegraphics[width=0.8\linewidth]{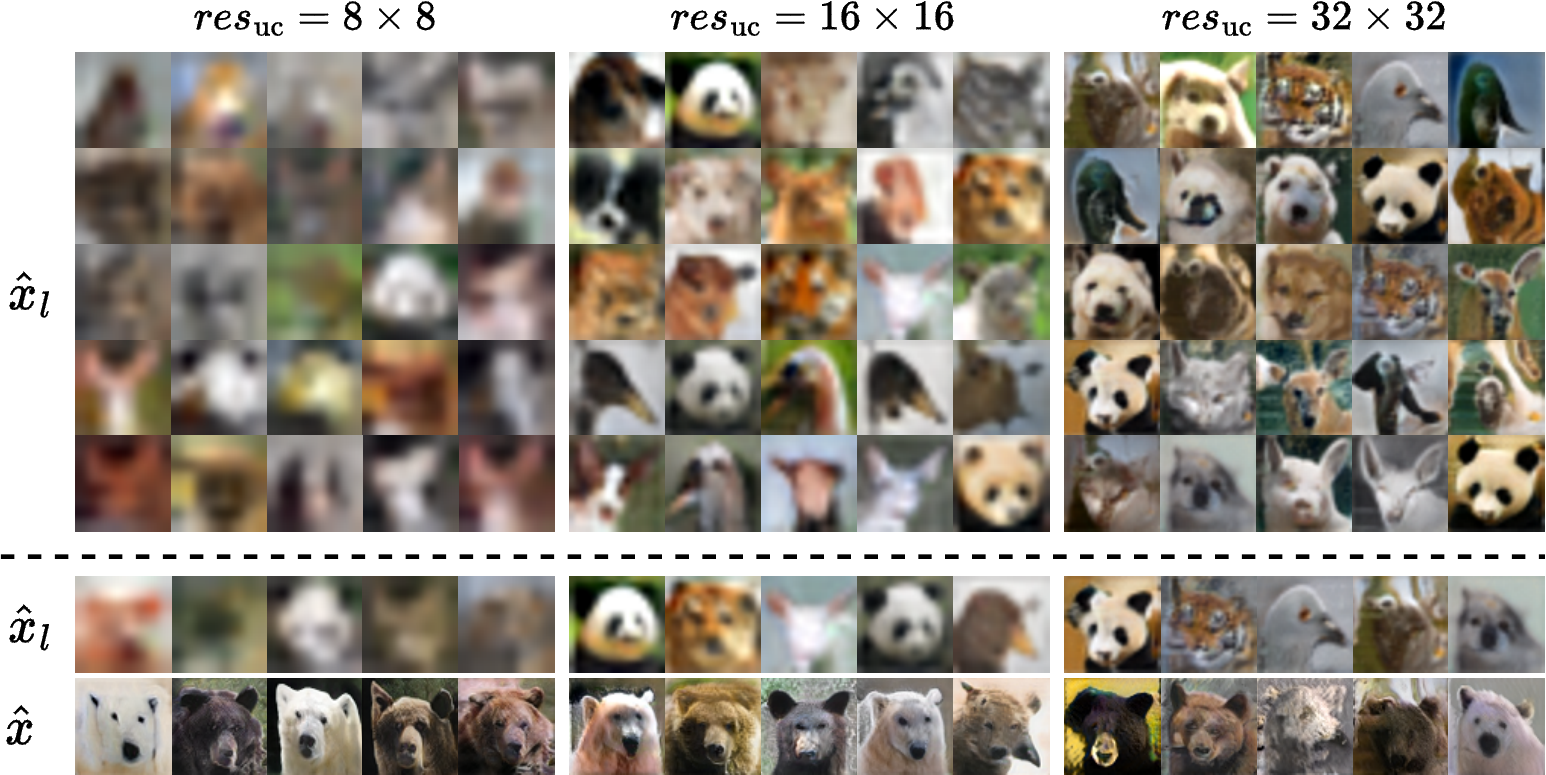}
  \caption{Visual comparison of the choice of different unconditional low-resolution ($res_{\text{uc}}$) in our proposed framework on the AnimalFaces-LT dataset. As ($res_{\text{uc}}$) increases, the unconditional low-resolution image $\hat{\vx}_l$ entails finer features (top). The low-resolution images are then used to generate class-conditional images at output resolution ($\hat{\vx}$) from the tail class \texttt{bear} (bottom).  The images are upsampled to the same size for better comparison. (best viewed in color)}
  \label{fig:ablation_res}
\end{figure}

To better understand the role of the intermediate low resolution selected during training, we show the unconditional low-res images $\hat{\vx}_l$ from the intermediate layers (see Fig.3 in the main paper) for different ablated resolutions in Fig. \ref{fig:ablation_res}. The low-res images are upsampled to the same size for better visual comparison. As $res{\text{uc}}$ increases, more details are introduced to the unconditionally trained intermediate images. For the AnimalFaces-LT dataset, resolution $32 \times 32$ already includes fine and definite details, making it challenging to generate (tail) class-specific changes in a single layer before reaching the output resolution of $64 \times 64$. We demonstrate this in the bottom of Fig. \ref{fig:ablation_res}  the final class-conditional images at $64 \times 64$ resolution from the tail class \texttt{bear} are shown. When $res_{\text{uc}}$ is higher, the final output changes minimally from the lower-resolution images $\hat{\vx}l$. On the other hand, at lower $res{\text{uc}}$, the level of change is higher.

\begin{table}[bht]
\caption{Ablation study on the choice of intermediate low resolution for unconditional training $res_{\text{uc}}$ for AnimalFaces-LT dataset ($64 \times 64$).}
\label{table:ablation_res}
\begin{center}
\begin{small}
\resizebox{.7\textwidth}{!}{%

\begin{tabular}{c|cc|cc}

\multirow{2}{*}{Unconditional Low Resolution ($res_{\text{uc}}$)} & 
\multirow{2}{*}{FID $\downarrow$} &
\multirow{2}{*}{FID-FS $\downarrow$}  &
{KID $\downarrow$} & 
{KID-FS  $\downarrow$} \\

{} & \multicolumn{2}{c|}{} & \multicolumn{2}{c}{$\times 1000$} \\
\hline \hline                    
     
\multirow{1}{*}{$\bb{8 \times 8}$} 
&  {$\bb{26.2}$}&{$\bb{48.4}$} & {$\bb{12.6}$}&{$\bb{19.6}$} \\

\hline 
\multirow{1}{*}{$\uu{16 \times 16}$} 
&  {$\uu{27.5}$}&{$\uu{50.3}$} & {$\uu{13.7}$}&{$\uu{20.8}$} \\

\hline 
\multirow{1}{*}{$32 \times 32$} 
& {${38.0}$}&{${64.9}$} & {${23.3}$}&{${34.3}$} \\

\end{tabular}
}
\end{small}
\end{center}
\end{table}

\textbf{Contribution of Unconditional Training Objective at Low Resolutions.} Another hyperparameter introduced by our method is the choice of unconditional training objective weight ($\lambda$) relative to the conditional one (see Eq. 2\&3 in the main paper). Since the results for $8 \times 8$ and $16 \times 16$ resolutions were comparable in Table. \ref{table:ablation_res}, we conducted an ablation study on $\lambda$ values for both resolutions. Table. \ref{table:ablation_lambda} shows the results of the ablation study on the contribution of different values for the unconditional training objective ($\lambda={0.01,0.1,1,10}$). We also considered the case when no unconditional training is added ($\lambda=0$), and only the lower resolution layers $\le res_{\text{uc}}$ do not receive class-conditional information, i.e., they are passed $\vw_z$ as the style vector instead of $\vw_{z,y}$ (see Figure 3 in the main paper). We find that $\lambda=1$ achieves the best performance for both resolutions. When $\lambda$ is too small ($0.01$) or too large ($10$), it disrupts the balance between the conditional and unconditional objectives, leading to performance degradation.

\begin{table}[ht]
\caption{\small Ablation on the choice of unconditional training objective weight ($\lambda$) at different low-resolutions $res_{\text{uc}}$ for AnimalFaces-LT dataset.}
\label{table:ablation_lambda}
\begin{center}
\begin{small}
\resizebox{0.86\textwidth}{!}{%

\begin{tabular}{c|c|cc|cc}

\multirow{2}{*}{$res_{\text{uc}}$} & 
\multirow{2}{*}{Unconditional Training Objective Weight ($\lambda$)} & 
\multirow{2}{*}{FID $\downarrow$} &
\multirow{2}{*}{FID-FS $\downarrow$}  &
{KID $\downarrow$} & 
{KID-FS  $\downarrow$} \\

{} & {} & \multicolumn{2}{c|}{} & \multicolumn{2}{c}{$\times 1000$} \\
\hline \hline                    

\multirow{5}{*}{$8 \times 8$} & {No Unconditional Training ($\lambda=0$)}
&  {${64.0}$}&{${94.2}$} & {${35.8}$}&{${45.4}$} \\
\cline{2-6}

{} & {$0.01$}
&  {${61.0}$}&{${99.9}$} & {${32.4}$}&{${50.0}$} \\

{} & {$\uu{0.1}$}
&  {${\uu{28.6}}$}&{$\uu{50.0}$} & {$\uu{13.6}$}&{$\uu{19.8}$} \\

{} & {$\bb{1}$}
&  {$\bb{26.2}$}&{$\bb{48.4}$} & {$\bb{12.6}$}&{$\bb{19.6}$} \\

{} & {$10$}
&  {${111.6}$}&{${145.8}$} & {${54.6}$}&{${66.4}$} \\

\hline \hline

\multirow{5}{*}{$16 \times 16$} & {No Unconditional Training ($\lambda=0$)}
&  {${58.3}$}&{${87.7}$} & {${28.5}$}&{${41.8}$} \\
\cline{2-6}

{} & {$0.01$}
&  {${64.7}$}&{${88.5}$} & {${30.7}$}&{${35.9}$} \\

{} & {$\uu{0.1}$}
&  {$\uu{31.4}$}&{$\uu{53.3}$} & {$\uu{15.7}$}&{$\uu{22.6}$} \\

{} & {$\bb{1}$}
&  {$\bb{27.5}$}&{$\bb{50.3}$} & {$\bb{13.7}$}&{$\bb{20.8}$} \\

{} & {$10$}
&  {${59.9}$}&{${111.0}$} & {${30.0}$}&{${53.8}$} \\

\end{tabular}
}
\end{small}
\end{center}
\end{table}

\textbf{Need for unconditional layers in the discriminator ($\mathcal{L}_{uc}$) and end-to-end joint training with both unconditional and conditional objectives.}
In addition to the unconditional layers in the generator, we have found that unconditional layers should be explicitly present in the discriminator. Table \ref{table:rebut_comb} provides the ablation results where the unconditional layers are removed from UTLO (i.e., w/o $\mathcal{L}_{uc}$). This shows significantly worse performance, indicating the need for \textit{explicit} unconditional discriminator on the low resolution. 

Further, to demonstrate how the proposed method differs from other training strategies that promote coarse-to-fine learning, e.g. progressive training~\cite{SauerS022StyleGAN-XL}, we carefully design experiments to compare our proposed method against progressive training. We follow the progressive strategy in StyleGAN-XL~\cite{SauerS022StyleGAN-XL}, starting with training a stem at a low resolution. After training the stem, the training of higher-resolution layers is followed. For a more comprehensive analysis, we experimented with two stems: an unconditional stem and a conditional one. 

Firstly, we observed the conditional stem exhibited mode collapse early in the training. For training the subsequent higher-resolution layers, we picked the best stem before the mode collapse. This training strategy yielded considerably worse results as shown in Table \ref{table:rebut_comb} (last row). Conversely, the unconditional low-resolution stem did not experience mode collapse. Indeed, Table \ref{table:rebut_comb} shows that using unconditional stem (second-to-last row) improved over the baseline. However, it was still significantly worse than UTLO, showing the benefit of our design: end-to-end joint training with both unconditional and conditional objectives.

\begin{table}[ht]
\caption{}
\label{table:rebut_comb}
\begin{center}
\begin{small}
\resizebox{0.7\linewidth}{!}{%

\begin{tabular}{l|cc|cc}

\multirow{2}{*}{Method} & 
\multirow{2}{*}{FID $\downarrow$} &
\multirow{2}{*}{FID-FS $\downarrow$}  &
{KID $\downarrow$} & 
{KID-FS  $\downarrow$} \\

{} & \multicolumn{2}{c|}{} & \multicolumn{2}{c}{$\times 1000$} \\
\hline \hline                    

\multirow{1}{*}{StyleGAN2-ADA~\cite{karras2020training}} 
&  {${51.4}$}&{$87.1$} & {${24.7}$}&{$35.9$} \\

\multirow{1}{*}{+ UTLO (Ours)} 
&  {$\bb{26.2}$}&{$\ul{48.4}$} & {$\ul{12.6}$}&{$\ul{19.6}$}\\

\hline
\multirow{1}{*}{+ UTLO w/o $\mathcal{L}_{uc}$} 
&  {${63.75}$}&{$87.35$} & {${35.08}$}&{$35.78$}\\

\hline
\multirow{1}{*}{+ Progressive~\cite{SauerS022StyleGAN-XL}: Uncond. Stem} 
&  {${37.73}$}&{$64.57$} & {${21.71}$}&{$31.53$} \\

\multirow{1}{*}{+ Progressive~\cite{SauerS022StyleGAN-XL}: Cond. Stem} 
&  {${98.35}$}&{$150.86$} & {${73.97}$}&{$102.20$} 

\end{tabular}
}
\end{small}
\end{center}
\end{table}

\section{Knowledge Sharing Analysis}\label{supp:knowledge_sharing}

In Figure 5 of the main paper, we presented conditionally generated images from our method trained on the CIFAR10-LT ($\rho=100$) dataset, demonstrating knowledge sharing among the head and tail categories using a shared unconditional low-resolution image. To quantify the similarity among images that share the same input latent code but have different class-condition labels, we use the Learned Perceptual Image Patch Similarity (LPIPS) metric \cite{zhang2018unreasonable}.
We first generate 1,000 random noises from different seeds. Given each sampled noise $z$, we generate images from all 10 classes of the CIFAR10-LT ($\rho=100$) dataset, $c\in \{0,\dots,9\}$, including both head and tail classes. We then calculate the LPIPS among all class pairs. Table. \ref{table:lpips} reports the average LPIPS score obtained from the baselines and our proposed method.

\begin{table}[bht]
\caption{\small Comparing Avg. LPIPS among all class pairs. Given a fixed noise input, we generate images from all classes (including both head and tail). We then quantify the similarities in terms of LPIPS metric among all class pairs and compare our method against baselines.}
\label{table:lpips}
\begin{center}
\begin{small}
\resizebox{0.681\textwidth}{!}{%

\begin{tabular}{c||c|c|c}

{Methods} & 
{StyleGAN2-ADA \cite{karras2020training}} &
{+ GSR \cite{rangwani2022improving} }  &
{+ UTLO (Ours)} \\

\hline               

{Avg. LPIPS $\times 1000$} & {$12.31$}&{$12.37$}&{$11.20$} \\

\end{tabular}
}
\end{small}
\end{center}
\end{table}

\begin{figure}[bht]
  \centering
  \includegraphics[width=1\linewidth]{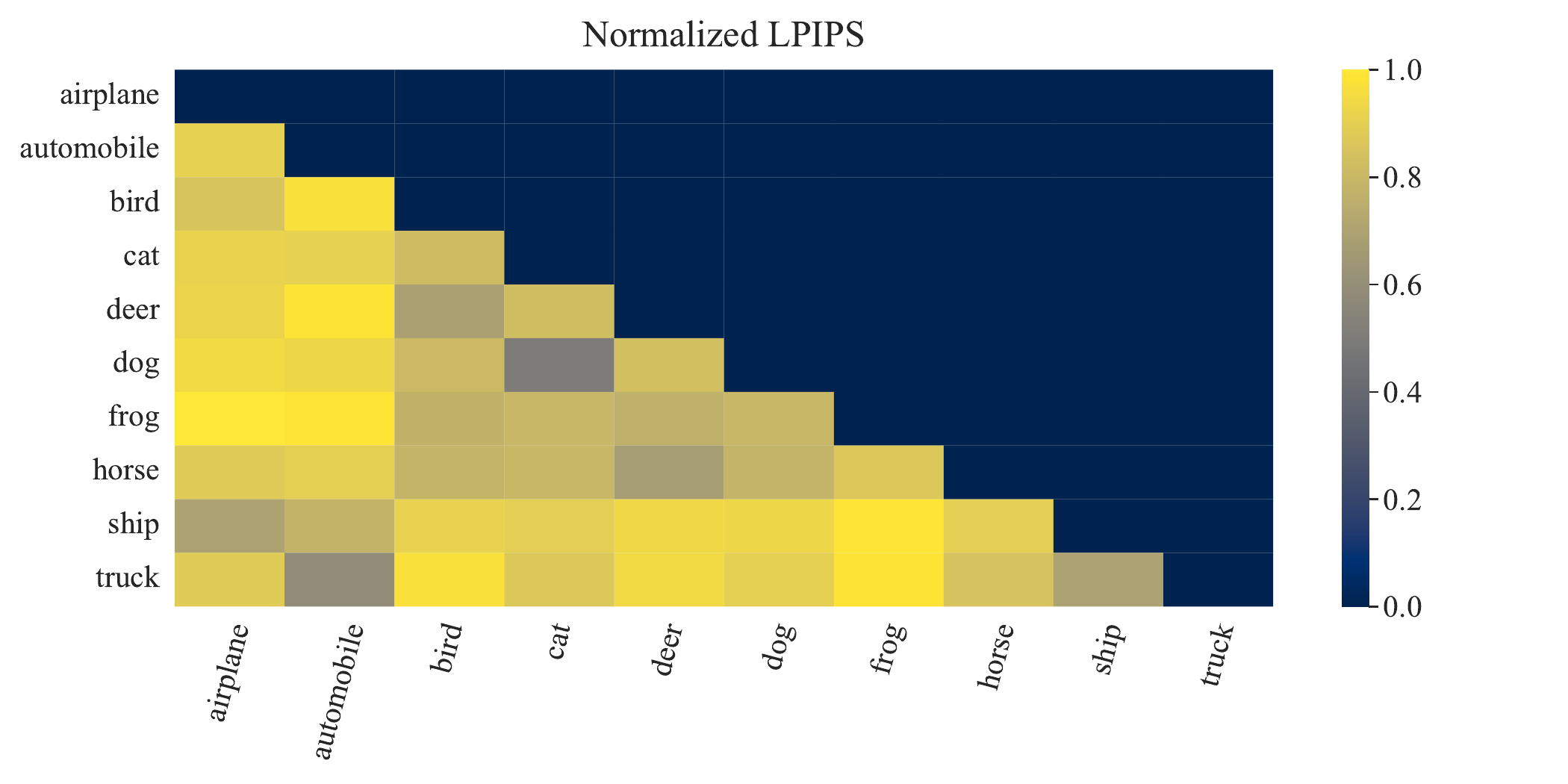}
   \caption{Normalized LPIPS among all class pairs of CIFAR10-LT dataset using our proposed method.  We generate images from all classes using the same latent codes and calculate the LPIPS among all class pairs. The obtained results are normalized from 0 to 1.We observe that tail classes such as \texttt{truck}, \texttt{ship}, and \texttt{frog} share the highest similarities with head classes \texttt{automobile}, \texttt{airplane}, and \texttt{bird}, respectively. The lower LPIPS values indicate higher similarities between the generated images. (best viewed in color)}
   \label{fig:lpips}
\end{figure}

The results show that generated images from all classes (head and tail) in UTLO, which promotes knowledge sharing, exhibit higher similarities (lower LPIPS) compared to the baselines that do not incorporate means of knowledge-sharing. To investigate which head and tail classes share the most patch similarities, we plotted the average LPIPS for individual class pairs in Fig. \ref{fig:lpips}. As somewhat expected,  It is observed that tail classes such as \texttt{truck}, \texttt{ship}, and \texttt{frog} share the highest similarities with head classes \texttt{automobile}, \texttt{airplane}, and \texttt{bird}, respectively. This is intuitive as these class pairs have common attributes such as \textit{blue/green backgrounds, wheels, etc.} Visual examples of this can also be found in Figure 5 of the main paper and Fig. \ref{fig:cf10++}.

\section{Long-tail v.s. Limited Data Regime}\label{supp:E}
Previous work on training GANs under limited data \cite{shahbazi2022collapse,karras2020training,zhao2020differentiable} has shown that the quality of generated images degrades as the dataset size decreases. In long-tailed datasets, on the other hand, there is an additional challenge of data imbalance across classes. To better understand their individual effects on training cGANs, we devise a setup in which we create a balanced dataset with the same size as the long-tail one. More concretely, given a long-tailed dataset with a total of $n$ training images (across head and tail classes), we create a new balanced dataset of size $n$ in which each class $c \in {1,\dots, C}$ contains $n/C$ images.

\begin{table}[bht]
\caption{Effect of (balanced) limited data v.s. long-tailed data in training cGANs. This table compares the quality of the generated images for different dataset sizes and distributions on CIFAR10 dataset. Baseline StyleGAN2-ADA  is used \textbf{without} GSR or the proposed UTLO.
}
\label{table:uniform}
\begin{center}
\begin{small}
\resizebox{0.8\textwidth}{!}{%

\begin{tabular}{c|c|cc|cc}

\multirow{2}{*}{\# Train Images} & 
\multirow{2}{*}{Data Distribution} & 
\multirow{2}{*}{FID $\downarrow$} &
\multirow{2}{*}{FID-FS $\downarrow$}  &
{KID $\downarrow$} & 
{KID-FS  $\downarrow$} \\

{} & {} & \multicolumn{2}{c|}{} & \multicolumn{2}{c}{$\times 1000$} \\
\hline \hline                    

\multirow{1}{*}{50,000} & {Full Dataset (Balanced)}    
&  {$2.5$}&{$3.3$}&{$0.4$}&{$0.4$} \\ 
 
 \hline                    
 
\multirow{2}{*}{13,996} & {Long-tail ($\rho=50$)}    
&  {$6.5$}&{$21.4$}&{$2.4$}&{$9.0$} \\ 
{} & {Limited Data (Balanced)}
&  {${3.9}$}&{${4.6}$} & {${1.0}$}&{${0.8}$} \\

\hline              

\multirow{2}{*}{12,406} & {Long-tail ($\rho=100$)}    
&  {$9.0$}&{$24.2$} & {$4.0$}&{$9.7$} \\
{} & {Limited Data (Balanced)}
&  {${4.5}$}&{${5.7}$} & {${1.3}$}&{${1.3}$} \\

\end{tabular}
}
\end{small}
\end{center}
\end{table}

To analyze the effects of dataset size v.s. the distribution of the classes in the dataset, we trained the baseline StyleGAN2-ADA \cite{karras2020training} on different setups of CIFAR10. Table. \ref{table:uniform} compares the quality of the generated images in terms of GAN metrics for each setup. We observe that in the balanced setup, the model achieves better scores compared to the long-tail setup. Moreover, we see that the performance gap between the full and few-shot metrics is noticeably larger in the long-tail setup.
We suspect that the performance gap between the full and few-shot metrics in the balanced setup might be due to the difficulty of the selected few-shot classes or the size of the real data used in calculating the metric. A smaller gap between the KID and KID-FS, which is unbiased in design, supports the latter.\\

\section{Implementation Details, and Choice of Hyperparameters}\label{supp:impl}
The baselines used in our experiments cover different designs in generator, discriminator, and data augmentation pipelines. We provide the implementation details in the following.

\begin{itemize}
    \item \textbf{StyleGAN2 with Adaptive Data Augmentation (ADA)} \cite{karras2020training}: We use the official PyTorch implementation \footnote{https://github.com/NVlabs/stylegan2-ada-pytorch} in our experiments. On CIFAR-LT datasets, we used the \texttt{cifar} configuration. For the rest of the datasets, we use the \texttt{auto} configuration. Adding transitional training \cite{shahbazi2022collapse}, we used the official implementation \footnote{https://github.com/mshahbazi72/transitional-cGAN} provided by the authors.\\
    
    \item \textbf{Projected GAN (PGAN)} \cite{sauer2021projected}: We use the projected discriminator with both the \textit{FastGAN} \cite{liu2021towards} and StyelGAN2 \cite{stylegan2Karras} generator backbones in our experiments. For the data augmentation, \textit{Differentiable Augmentation (DA)} \cite{zhao2020differentiable} is used. The official PyTorch implementation is provided by the authors \footnote{https://github.com/autonomousvision/projected-gan}. Default hyperparameters are used. Note, the authors of PGAN provided a liter version of FastGAN which gets to similar performance as the original FastGAN. We use the lite version in our experiments.\\
    
    \item \textbf{Group Spectral Regularization (GSR)} \cite{rangwani2022improving}: We added the GSR implementation provided by the authors \footnote{https://github.com/val-iisc/gSRGAN} to both the StyleGAN2+ADA and PGAN+DA repositories. We used the default choice of hyperparameters in all experiments. \\

    \item \textbf{Noisy Twins} \cite{rangwani2023noisytwins}: We use the official implementation provided by the authors \footnote{https://github.com/val-iisc/NoisyTwins} and also added it to the PGAN+DA repository. We used the default choice of hyperparameters in all experiments. \\

    \item \textbf{Unconditional Training at Lower Resolutions (UTLO)}: We used the default training configuration of the baselines. Our method introduces two new hyperparameters: the intermediate low resolution and $\lambda$, the weight between the conditional and unconditional loss terms (see ablation in Sec. \ref{supp:ablation}). In general, we found intermediate low resolution of $8 \times 8$ and $\lambda=\{0.1,1\}$ to be generally applicable across all datasets and resolutions, attaining competitive performance.
    
    For the reported results in the paper, we use $8 \times 8$ as the intermediate low-resolution across all datasets, except for the Flowers-LT dataset where $16 \times 16$ resolution shows slightly better performance. Choosing $\lambda$, on CIFAR100-LT and Flowers-LT which contain very few samples in the tail classes, we found $\lambda=10$ performing the best. On CIFAR10-LT, $\lambda$ is set to 0.1. Across the rest of the datasets, we use $\lambda =1$. 
\end{itemize}

\section{Additional Evaluation Results}\label{supp:G}

\noindent\textbf{Naturally Imbalanced Datasets: iNaturalist2019 and Flowers-LT}
We evaluate our proposed method on datasets that are naturally imbalanced. This covers the iNaturalist2019 and Flowers-LT datasets which we train at $64 \times 64$ and $128 \times 128$ resolutions, respectively. Different from previous experiments, we use ProjectedGAN (StyleGAN2) + DA \cite{sauer2021projected} as the baseline here. Table. \ref{table:inat_fl} shows the quantitative evaluation results. %
The results across both datasets consolidate our findings and validate the effectiveness of our proposed method.

\begin{figure}[hbt]
  \centering
  \includegraphics[width=0.65\linewidth]{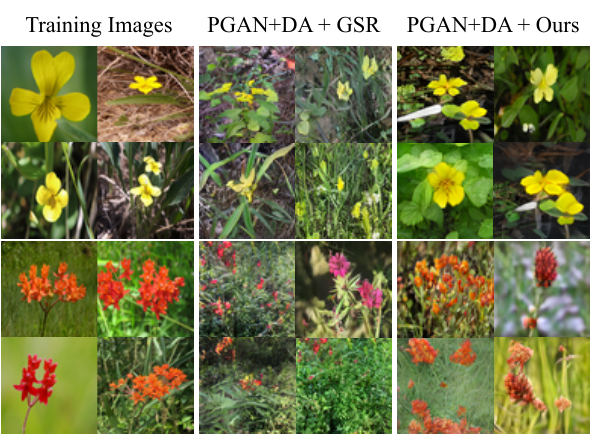}

   \caption{Examples of training and generated images contrasting our method against the baseline on two different tail classes of the iNaturalist2019 dataset at $64 \times 64$ resolution. Each row depicts a tail class.}
   \label{fig:inat}
   \vskip -0.1in
\end{figure}

Additionally, we show visual examples from the training and generated images from two different tail classes in the iNaturalist2019 dataset in Fig. \ref{fig:inat}. It can be seen that the images learned from our method (UTLO) tend to generate more diverse and higher-quality images for the tail classes. Generated images from the tail classes of the Flowers-LT dataset are shown in Fig. \ref{fig:fl+}.

\begin{table*}[hbt]

\caption{\small Comparing the proposed method against the baseline across two naturally imbalanced datasets: Flowers-LT ($128 \times 128$ resolution) and iNaturalist2019 ($64 \times 64$ resolution).}
\label{table:inat_fl}
\begin{center}
\begin{small}
\resizebox{0.9\textwidth}{!}{%

\begin{tabular}{l|cccc|cccc}

{Dataset} & \multicolumn{4}{c}{Flowers-LT} & \multicolumn{4}{c}{iNaturalist2019}\\
\hline\hline

\multirow{2}{*}{Metrics} & 
\multirow{2}{*}{FID $\downarrow$} &
\multirow{2}{*}{FID-FS $\downarrow$}  &
{KID $\downarrow$} & 
{KID-FS  $\downarrow$} &
\multirow{2}{*}{FID $\downarrow$} &
\multirow{2}{*}{FID-FS $\downarrow$}  &
{KID $\downarrow$} & 
{KID-FS  $\downarrow$} \\

{} & \multicolumn{2}{c}{} & \multicolumn{2}{c|}{$\times 1000$} &  \multicolumn{2}{c}{} &  \multicolumn{2}{c}{$\times 1000$}\\

\hline \hline
{PGAN (StyleGAN2) + DA~\cite{sauer2021projected}}
&  {$9.8$}&{$21.6$}&{$2.4$}&{$2.9$} & 
{$3.6$}&{$11.4$}&{$0.53$}&{$1.08$}\\

{+ GSR\cite{rangwani2022improving}}              
&  {$8.2$}&{$17.9$}&{$1.1$}&{$\bb{1.7}$} & 
 {$3.5$}&{$11.1$}&{$0.51$}&{$0.99$} \\ 

{+ NoisyTwins\cite{rangwani2023noisytwins}}                    
&  {$6.7$}&{$\bb{15.3}$}&{$\bb{0.9}$}&{$1.9$} & {$3.0$}&{$10.6$}&{$0.45$}&{$0.72$} \\ 

{+ UTLO (Ours)}                    
& {$\bb{6.6}$}&{$15.4$}&{$\bb{0.9}$}&{$1.8$} &  

{$\bb{2.8}$}&{$\bb{10.1}$}&{$\bb{0.41}$}&{$\bb{0.60}$} \\  

\end{tabular}
}
\end{small}
\end{center}
\vskip -0.2in
\end{table*}

\textbf{Combining our method with GSR and WS}
Although we provided a direct comparison of UTLO against GSR \cite{rangwani2022improving} and weighted sampling (WS), they can be integrated with our proposed training framework. In Table. \ref{table:res+}, we present quantitative results on AnimalFaces-LT when adding GSR, NoisyTwin, and WS (see Sec. \ref{supp:ws}) to UTLO. We use WS with $\beta=0.35$. Combining UTLO
with regularization methods (GSR and NoisyTwins) and weighted sampling (WS) didn’t show clear improvements over UTLO alone. However, it resulted in significant improvements over GSR and NoisyTwins individually. \\

\begin{table}[t]
\caption{Integrating GSR and weighted sampling (WS) to our proposed method on the AnimalFaces-LT dataset. The obtained results from adding GR and WS do not show any noticeable improvement over solely using UTLO.}
\label{table:res+}
\begin{center}
\begin{small}
\resizebox{.6\textwidth}{!}{%

\begin{tabular}{l|cc|cc}

\multirow{2}{*}{Method} & 
\multirow{2}{*}{FID $\downarrow$} &
\multirow{2}{*}{FID-FS $\downarrow$}  &
{KID $\downarrow$} & 
{KID-FS  $\downarrow$} \\

{} & \multicolumn{2}{c|}{} & \multicolumn{2}{c}{$\times 1000$} \\
\hline \hline                    

\multirow{1}{*}{StyleGAN2-ADA} 
&  {${51.4}$}&{$87.1$} & {${24.7}$}&{$35.9$} \\

\multirow{1}{*}{+ UTLO (Ours)} 
&  {${26.2}$}&{${48.4}$} & {${12.6}$}&{${19.6}$}\\

\hline 
\multirow{1}{*}{+ GSR~\cite{rangwani2022improving}}              
&  {${39.2}$}&{$67.2$} & {${21.2}$}&{$32.7$}\\

\multirow{1}{*}{+ UTLO + GSR} 
&  {$\ul{26.9}$}&{$\ul{47.8}$} & {$\bb{12.2}$}&{$\ul{19.3}$}\\

\hline
\multirow{1}{*}{+ NoisyTwins~\cite{rangwani2023noisytwins}}
&  {${29.4}$}&{$50.2$} & {${16.7}$}&{$21.2$}\\

\multirow{1}{*}{+ UTLO  + NoisyTwins} 
&  {$\ul{26.6}$}&{$\bb{48.1}$} & {$\ul{12.9}$}&{$\bb{19.0}$}\\

\hline
\multirow{1}{*}{StyleGAN2-ADA + UTLO  + WS} 
&  {${27.7}$}&{$47.4$} & {${13.4}$}&{$18.5$}\\

\end{tabular}
}
\end{small}
\end{center}
\end{table}

\textbf{Comparison against Unconditional Training}
We compare unconditional and conditional training on the AnimalFaces-LT dataset and present the results in Table. \ref{table:res++}. The unconditional model generates samples that follow the training distribution, which is mainly dominated by head classes. This bias favors the unconditional baseline in terms of the FID and KID metrics, which do not consider the skewness in the data distribution. We recommend including FID-FS and KID-FS metrics when evaluating GANs on imbalanced datasets.\\

\begin{table}[t]
\caption{Comparing unconditional and conditional baselines on AnimalFacs-LT dataset. The Unconditional baseline generates samples that track the training distribution which is mainly dominated by head classes. This favors it in terms of FID and KID metrics, which do not consider the skewness in the data distribution. We suggest FID-FS and KID-FS metrics should be incorporated when evaluating GANs over imbalanced datasets. 
}
\label{table:res++}
\begin{center}
\begin{small}
\resizebox{.7\textwidth}{!}{%

\begin{tabular}{l|cc|cc}

\multirow{2}{*}{Method} & 
\multirow{2}{*}{FID $\downarrow$} &
\multirow{2}{*}{FID-FS $\downarrow$}  &
{KID $\downarrow$} & 
{KID-FS  $\downarrow$} \\

{} & \multicolumn{2}{c|}{} & \multicolumn{2}{c}{$\times 1000$} \\
\hline \hline                    

\multirow{1}{*}{StyleGAN2-ADA Unconditional} 
&  {${39.4}$}&{$104.1$} & {${17.3}$}&{$27.6$} \\

\hline 
\multirow{1}{*}{StyleGAN2-ADA Conditional} 
&  {${51.4}$}&{$87.1$} & {${24.7}$}&{$35.9$} \\

\hline \hline
\multirow{1}{*}{StyleGAN2-ADA Conditional + UTLO (Ours)} 
&  {${26.2}$}&{${48.4}$} & {${12.6}$}&{${19.6}$}\\

\end{tabular}
}
\end{small}
\end{center}
\end{table}

\section{Additional Visual Comparison}\label{supp:H}
\begin{figure}[bht]
  \centering
  \includegraphics[width=0.99\linewidth]{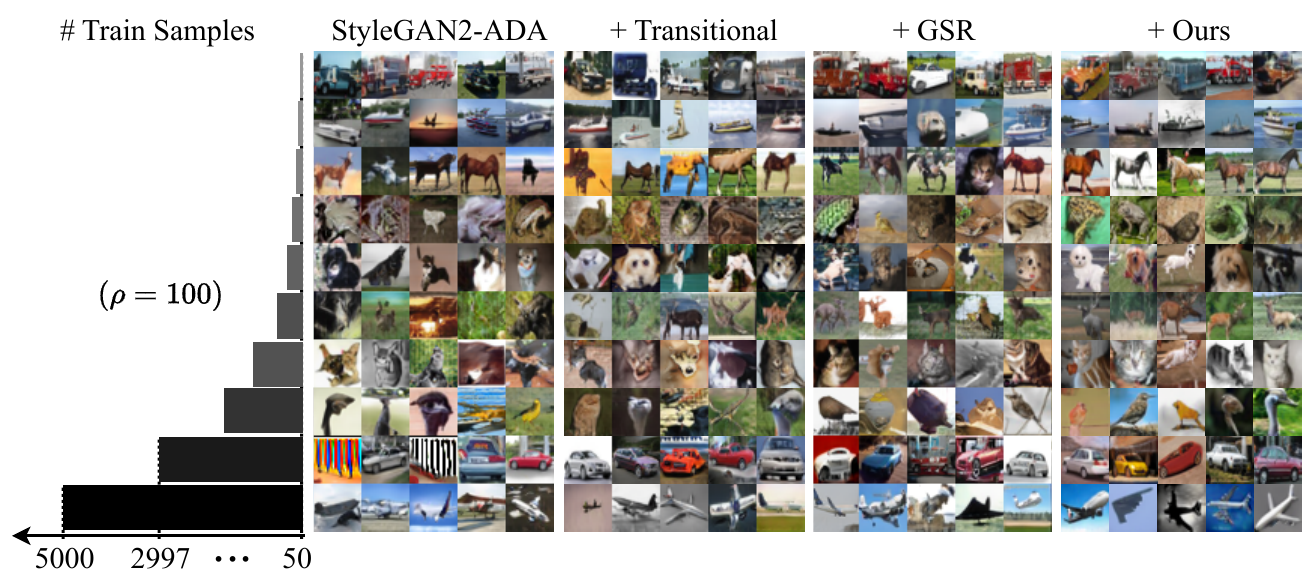}
   \caption{Qualitative comparison of the generated images from all classes in the CIFAR10-LT dataset ($\rho=100$). There are only 50 training samples present in the rarest tail class \texttt{truck} (top row) while the most populated head class \texttt{airplane} (bottom row) has 5,000 training samples.}
   \label{fig:cf10+}
\end{figure}

\begin{figure}[bht]
  \centering
  \includegraphics[width=0.5\linewidth]{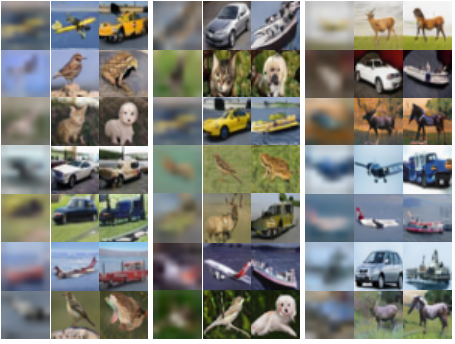}
   \caption{Additional examples of knowledge sharing from the head to the tail classes in
CIFAR10-LT dataset using our proposed UTLO framework. The conditional images generated from the head (middle
columns) and tail (right columns) classes share and are built on top
of the same low-resolution (unconditional) images (left columns).
Low-resolution images (8 × 8) are upsampled to that of CIFAR10-
LT (32 × 32) for improved visualization. (best viewed in color.)}
   \label{fig:cf10++}
\end{figure}

\begin{figure}[bht]
  \centering
  \includegraphics[width=0.65\linewidth]{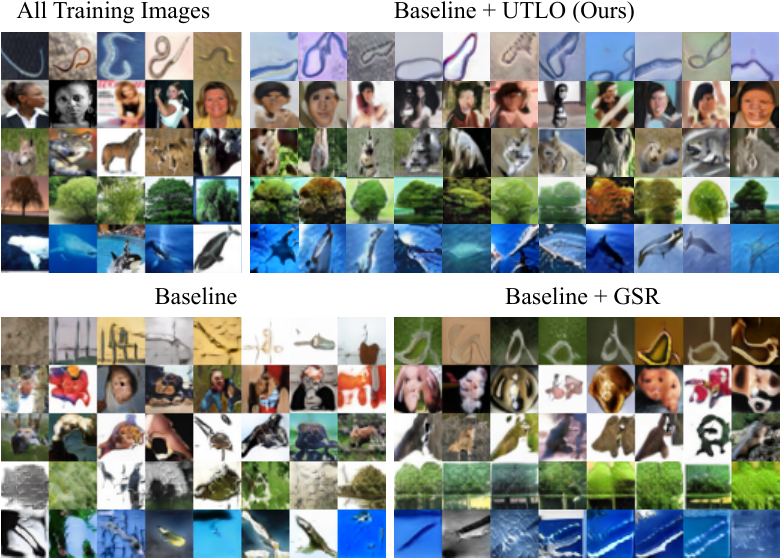}
   \caption{Comparing the generated images from our proposed method against the baseline across classes with only 5 training instances in the CIFAR100-LT dataset ($\rho=100$). The baseline used is StyleGAN2-ADA. Training images are shown in the top-left corner.}
   \label{fig:cf100+}
\end{figure}

\begin{figure}[bht]
  \centering
  \includegraphics[width=0.8\linewidth]{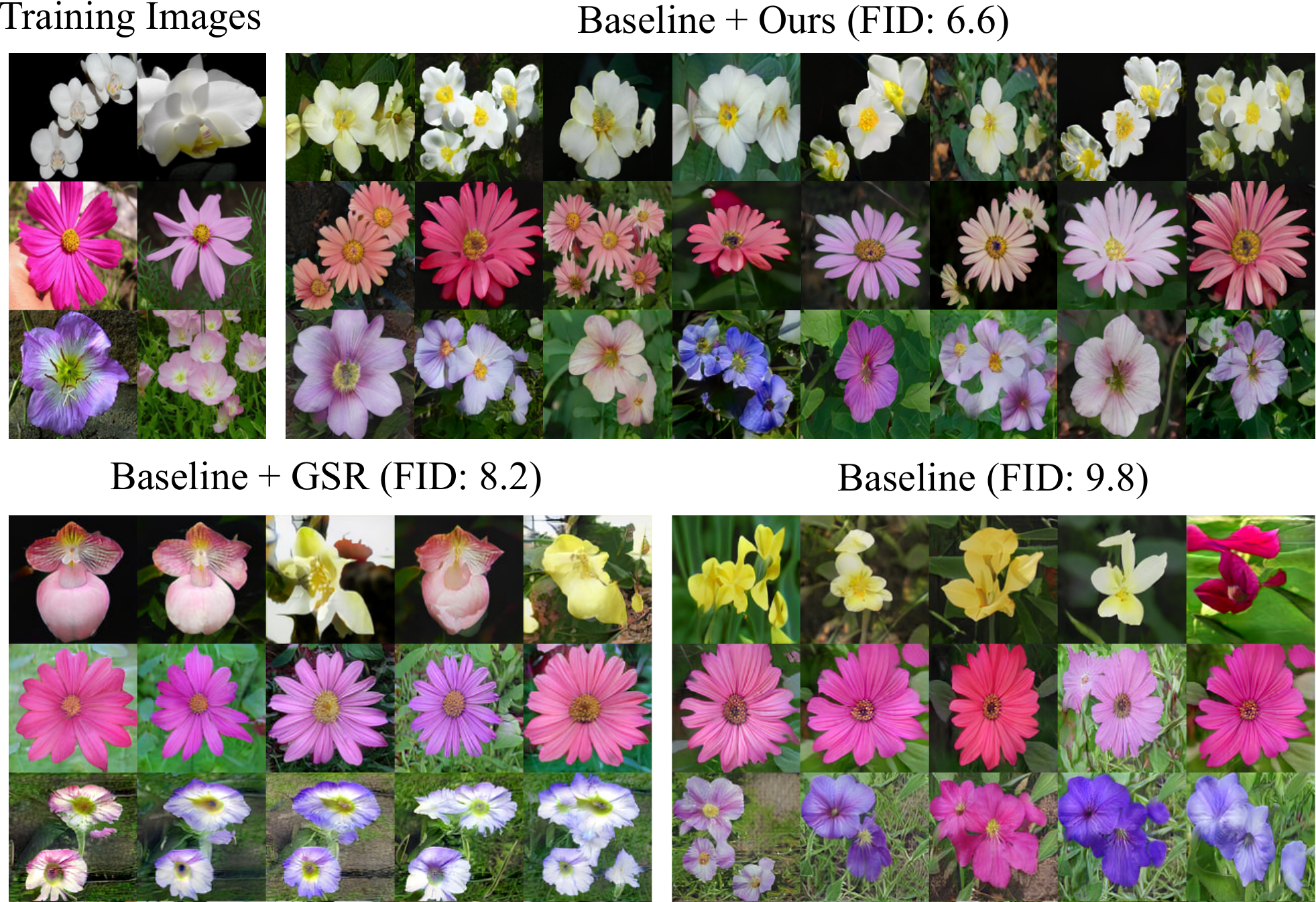}
   \caption{Additional visual examples when generating images from the rarest tail classes in the Flowers-LT with only 2 training images. Our proposed approach allows for a diverse set of features such as backgrounds, colors, poses, object layouts, etc. to be infused into the tail classes with very few training images. Training images are shown in the top-left corner. ProjectedGAN (StyleGAN2) + DA is used as the baseline. }
   \label{fig:fl+}
\end{figure}

\begin{figure}[bht]
  \centering
  \includegraphics[width=0.7\linewidth]{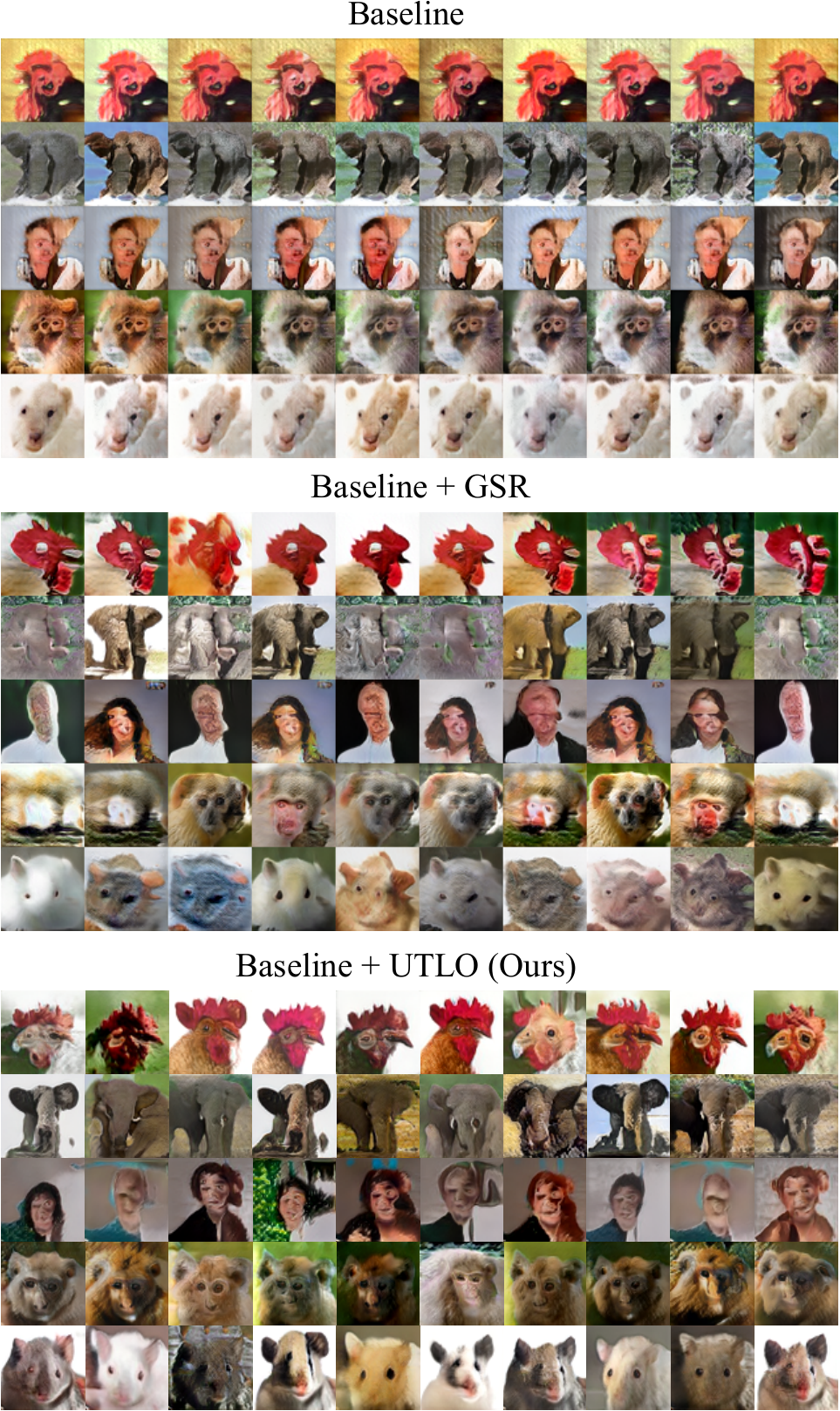}
   \caption{Generated images from 5 tail classes with the least number of training images in the AnimalFaces-LT dataset. While the diversity of the generated images is limited by baselines, UTLO learns a set of more diverse images with very few training images. StyleGAN2-ADA is used as the baseline. }
   \label{fig:af+}
\end{figure}

In Fig. \ref{fig:cf10+}-\ref{fig:af+}, we provide additional visual examples from our proposed method and compare them against baselines. Fig. \ref{fig:cf10+} presents a comparison of generated images from all classes in the CIFAR10-LT dataset ($\rho=100$). The data imbalance curve is shown in this figure where there are only 50 training samples present in the rarest tail class \texttt{truck} (top row) while the most populated head class \texttt{airplane} (bottom row) has 5,000 training samples. Fig. \ref{fig:cf10++} shows additional examples of knowledge sharing from the head to the tail classes in the CIFAR10-LT dataset using our proposed UTLO framework. The conditional images generated from the head (middle columns) and tail (right columns) classes share and are built on top of the same low-resolution (unconditional) images (left columns).\\

Fig. \ref{fig:cf100+} compares the generated images from our proposed method against the baseline across classes with \textit{only 5} training instances (shown in the top-left corner) in the CIFAR100-LT dataset ($\rho=100$). We use StyleGAN2-ADA as the baseline in the CIFAR100-LT experiments. Further, we present additional visual comparisons of generated images from the rarest tail classes of the Flowers-LT with \textit{only 2} training images (shown in the top-left corner) in Fig. \ref{fig:fl+}. Our proposed approach enables a diverse set of features, such as backgrounds, colors, poses, and object layouts, to be infused into the tail classes with very few training images. We use ProjectedGAN (StyleGAN2) + DA as the baseline. Finally, we showcase the generated images from the 5 tail classes with the least number of training images in the AnimalFaces-LT dataset. While the diversity of the generated images is limited by the baselines, UTLO learns a more diverse set of images with very few training images. StyleGAN2-ADA serves as the baseline for the AnimalFaces-LT experiments.\\

\clearpage

\end{document}